\documentclass[]{fairmeta}
\title{Embodied AI Agents: Modeling the World}

\author{Pascale Fung}
\author{Yoram Bachrach}
\author{Asli Celikyilmaz}
\author{Kamalika Chaudhuri}
\author{Delong Chen}
\author{Willy Chung}
\author{Emmanuel Dupoux}
\author{Hongyu Gong}
\author{Hervé Jégou}
\author{Alessandro Lazaric}
\author{Arjun Majumdar}
\author{Andrea Madotto}
\author{Franziska Meier}
\author{Florian Metze}
\author{Louis-Philippe Morency}
\author{Théo Moutakanni}
\author{Juan Pino}
\author{Basile Terver}
\author{Joseph Tighe}
\author{Paden Tomasello}
\author{Jitendra Malik}

\affiliation{Meta AI Research}

\renewcommand{\paragraph}[1]{\medskip\noindent {\bf #1}\quad}

\abstract{
This paper describes our research on AI agents embodied in visual, virtual or physical forms, enabling them to interact more naturally with both users and their environments. These agents, which include virtual avatars, wearable devices, and robots, are designed to perceive, learn and act within their surroundings, which makes them more similar to how humans learn and interact with the environments as compared to disembodied agents. We propose that the development of world models is central to reasoning and planning of embodied AI agents, allowing these agents to understand and predict their environment, to understand user intentions and social contexts, thereby enhancing their ability to perform complex tasks autonomously. World modeling encompasses the integration of multimodal perception, planning through reasoning for action and control, and memory to create a comprehensive understanding of the physical world. Beyond the physical world, we also propose to learn the mental world model of users to enable better human-agent collaboration. 

Virtual embodied agents are transforming fields such as therapy and entertainment by providing emotionally intelligent interactions. Wearable agents, for example on AI glasses, can potentially offer real-time assistance and personalized experiences, while robotic agents are poised to address labor shortages and perform tasks in unstructured environments. In addition to the technical challenges of embodied AI agents and our approach to solve them, we also outline the importance of ethical considerations, particularly concerning privacy and anthropomorphism, as these agents become more integrated into daily life.

Future research directions include embodied AI learning, improving multi-agent collaboration and human-agent interactions, enhancing their social intelligence, and ensuring ethical practices in designing in them. By addressing these challenges, embodied AI agents hold the promise of transforming human-technology interaction, making it more intuitive and responsive to human needs. This paper provides an overview of the current state and future directions of our research on embodied AI agents, on the path to unlocking their transformative impact on our lives. 
}

\date{\today}
\correspondence{Pascale Fung at \email{pascalefung@meta.com}}

\begin{document}

\maketitle

\tableofcontents

\section{Introduction}
\label{sec:introduction}
Embodied AI agents are artificial intelligence systems that are instantiated in a visual, virtual, or physical form, enabling them to learn and interact with both the user and their physical or digital surroundings. These embodied AI systems require the ability to perceive and act within their environment in meaningful ways, thereby necessitating a deep understanding of the physical world in which they operate. In contrast, AI agents that exist solely as web-based entities without a visual form do not possess embodiment, whereas robots or drones that are operated via tele-command or pre-programmed instructions lack the autonomy and adaptability characteristic of true AI agents. The unique nature of wearable devices distinguishes them from other smart devices, as they integrate AI systems that can perceive the physical world and execute actions within it. This synergy between perception and action enables wearable agents to be embodied from the user's perspective, effectively blurring the lines between human and machine. To paraphrase the philosopher Maurice Merleau-Ponty, "I am not in my body, I am my body" \citep{MerleauPonty1945}, highlighting the idea that the body is not just a vessel for the mind, but rather an integral part of our existence. This concept is closely related to the idea of embodiment in AI, where the agent's embodiment and environment are seen as an integral part of its cognitive processes. Building upon these ideas, we propose a framework for embodied AI agents that incorporates a world modeling approach, allowing these systems to reason about and interact with their environment in a more sophisticated and human-like manner.

Embodiment serves two primary purposes in current AI and robotics: (1) Physical Interaction: It enables AI systems to interact with the physical world, either through direct action (e.g., robotic agents) or by providing awareness of their surroundings (e.g., wearable agents); and (2) Enhanced Human-Machine Interaction: Embodied agents have been shown to foster greater user trust, as demonstrated in studies \citep{winata2017nora,fung2018towards,shridhar2024art}. Moreover, a growing area of exploration is (3) the potential for embodied agents to learn and develop in a manner similar to humans, leveraging a rich sensory experience that mirrors our own. This third purpose holds promise for more intuitive and human-like learning processes \citep{dupoux2018cognitive,radosavovic2023robot}.

A fully autonomous and self-learning AI system,  that is capable of interacting with humans and the world around us, and assisting us in our personal and professional lives, has been a long quest in the annals of AI development.   From the first rule-based chatbot, to AI call centre assistants, to virtual assistants, AI assistants have evolved with expanded and improved capabilities in each iteration. The emergence of online AI agents is the latest milestone in this development. Meanwhile, the embodiment of AI has assumed different forms, from embodied conversational agents with avatars \citep{cassell2001embodied} to wearable devices \citep{alsuradi2024neuro},  robots \citep{mon2025embodied} and  humanoid robots \citep{cao2024ai}. Each of these embodiments targets a different category of tasks and applications. Each of them necessitates different capabilities while sharing some essential capabilities in common. 

Unlike previous generations of AI assistants, AI agents are more autonomous. They can carry out tasks with multiple steps by figuring out on their own what steps they need to take, what external resources they need, and which other agents it needs to collaborate with, all the while understanding what the user needs either explicitly from user queries or implicitly from context. Embodied agents also need to plan and carry out actions, either on behalf of the user, or in assisting the user in their own actions. This requires a sophisticated level of reasoning and planning. To be able to perceive the world and then plan out actions is also known as world modeling \citep{PathAMI}. 

In addition, an agent should be able to converse with the user if it needs further clarification or confirmation, and if the context changes. In the future, agents can interact with each other and with multiple users. This human-agent interaction should be expressive, socially and contextually sensitive. In other words, agents need to understand the “mental world model” of the user. To support physical and mental world modeling, reasoning and planning, it is necessary for embodied agents to have short term and long term memory. 

\begin{figure}
    \centering
    \includegraphics[width=1\linewidth]{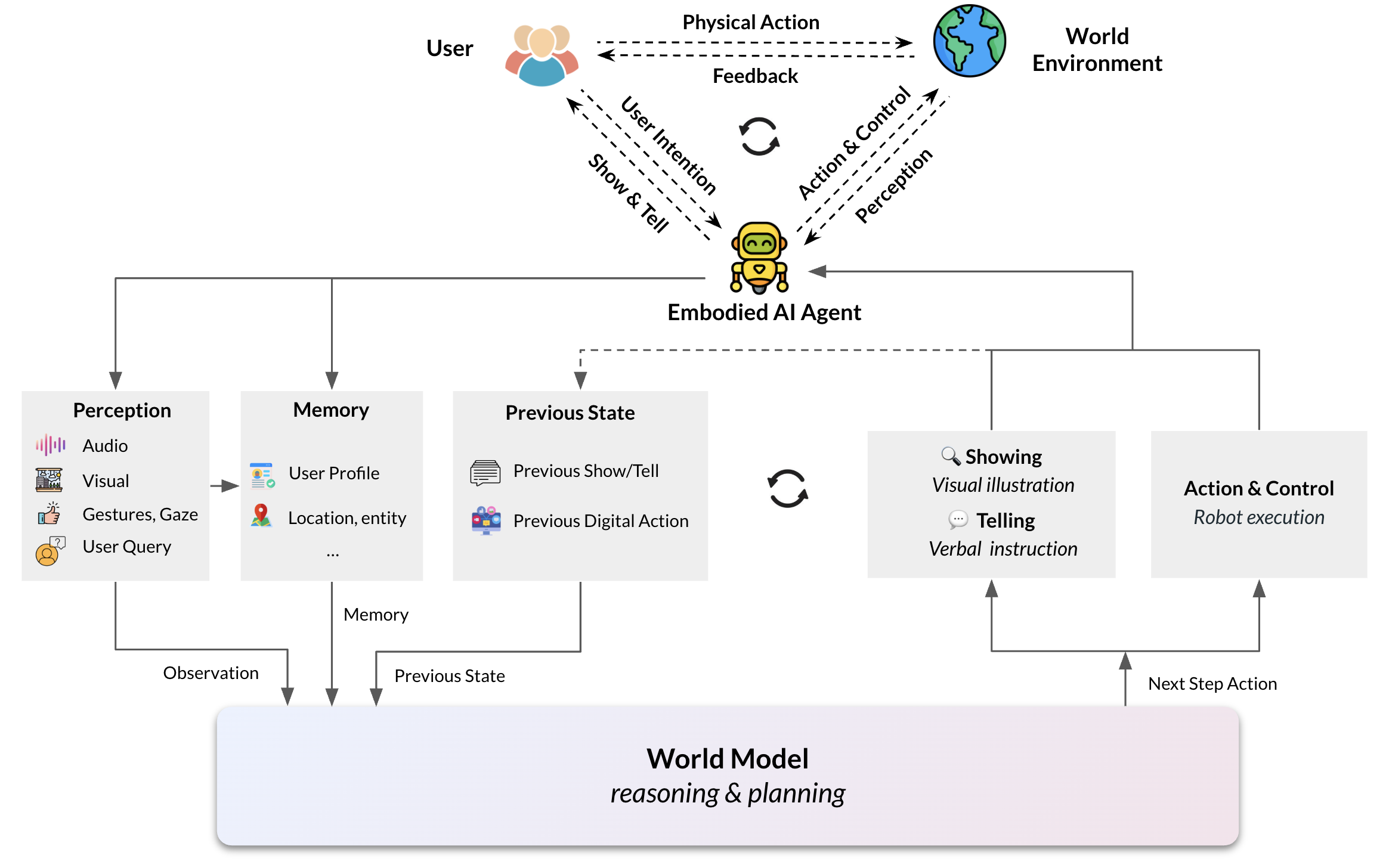}
    \caption{Overview of the Embodied AI Agent architecture, with the interaction loop between user, world and agent. The world model is the core component responsible for planning and reasoning.}
    \label{fig:embodied_ai_agent_archittecture}
\end{figure}

The evolution of AI assistants into AI agents today has been driven by the advancement in large language models and vision-language models. Developers have been prompting LLMs and VLMs to build embodied agents with avatars \citep{cherakara2023furchat}, on smart glasses, VR devices \citep{pan2024ellma}, and on robots \cite{brohan2023rt} and humanoid robots \citep{tong2024advancements}.  Large Language Models have been fine-tuned on conversational data, and also with RLHF to follow human preferences. They show both remarkable natural language understanding and generation abilities and, more importantly, showed zero-shot ability to carry out multiple tasks without any explicit design or training for such tasks. With millions of collective users of various LLMs in the market, the initial excitement of being able to talk to a chatbot about anything turned quickly into the expectation that they can assist humans in their jobs and improve their lives in any task with just a prompt. Recent announcements of online AI agents from different industry and academic labs are a result of this evolution. Smartglasses, such as Meta glass, allow a user to access AI agents, such as Meta Multimodal AI, based on what they see through the device camera and what they say through the device microphone, though not quite what they hear in their environment. The development of contextual AI is enabled by prompting LLMs and VLMs for perception as well as reasoning and planning \citep{erdogan2025plan}. VLMs can be instruction-tuned to produce planning step by step \cite{kim2024openvla}. Robotic actions can also be carried out following plans via prompting LLMs \citep{ahn2022can}. 

However, one fundamental flaw of generative models is the inefficiency in their model size. Generative models trained to predict the next token or pixel are excellent for creative tasks but they include too many textual or visual details while missing the essential information for reasoning and planning tasks. The capability for reasoning and planning is essential to AI agents. For both accuracy and efficiency in embodied AI, we propose a World Modeling approach that predicts actions and plans with reason from multimodal perception. 

In this paper, we first give an overview of different agent types and their applications, we then describe the framework proposed for embodied AI agents, namely world models, which includes perception, physical and mental world modeling, memory and action and control. We describe world models based on generative models as well as alternative predictive world models that are potentially more efficient and trustworthy. We then describe the capabilities and models for three types of embodied AI agents, namely (1) virtual embodied Agents; (2) wearable agents; and (3) robotic agents. In each section, we describe benchmarks for, as well as potential future research directions in each agent type.  Finally, we describe a future vision of embodied learning, and a Family of Agents that all work with each other with muti-agent collaboration.Last but not the least, we discuss two critical ethics issues with highly autonomous embodied AI agents, namely privacy and security as well as anthropomorphism.

\section{Embodied AI Agent Types and Applications}
\label{sec:agent_types_and_applications}
\subsection{Virtual Embodied Agents}
Virtual embodied agents (VEA) have emerged as a vital component in various conversational tasks, where the ability to convey emotions facilitate effective human-computer interactions. These agents can take on diverse forms, ranging from virtual 2D or 3D avatars to robotic androids with physical skins, human-like facial features, and motorized systems that enable the control of facial emotions and lip movements.

One of the most common applications of VEAs is in the field of AI therapy, where they are used to provide emotional support and companionship to individuals in need \citep{bickmore2005}. For instance, AI-powered chatbots like Woebot and Wysa have been designed to offer cognitive behavioral therapy and emotional support to users, leveraging the power of embodied conversational agents to create a more engaging and empathetic experience.

Beyond AI therapy, VEAs have also found applications in the realm of the Metaverse and in mixed reality. In virtual environments like Second Life, Horizon Worlds, and Sansar, VEAs can serve as guides, mentors, or even friends, providing users with a more immersive and interactive experience. VEAs can also help provide a richer background virtual environment, for example in the form of NPCs interacting with each other. These agents can be rendered to exhibit emotions, empathy, and social intelligence, making them an invaluable asset in virtual communities.

Another area where VEAs are being explored is in AI studio avatars. These avatars can be used in film, television, and video game production to create more realistic and engaging characters. By incorporating VEAs into the production process, creators can craft characters that are not only visually stunning but also emotionally intelligent, capable of conveying complex emotions and subtle nuances.

Furthermore, VEAs will play a crucial role in virtual and mixed reality environments to augment user’s experience with assistants, companions, and non-player characters in games. These avatars will be embodied and capable of perception and understanding of the virtual world and they will be able to control their body and its surroundings to accomplish tasks in collaboration with the user as well as other virtual agents. The UNICORNs team focuses on developing behavioral foundation models to control virtual embodied agents capable of solving complex tasks in a zero-shot fashion. In particular, the Meta Motivo model is trained to control a physics-based humanoid avatar to accomplish whole-body tasks prompted through poses to reach, motions to track, and objective functions to optimize.

In addition to these applications, VAEs have the potential to revolutionize various industries, including education, customer service, and healthcare. For example, VAEs can be used to create personalized learning experiences, providing students with adaptive feedback and emotional support. In customer service, ECAs can help resolve issues more efficiently, offering empathetic and human-like interactions that improve customer satisfaction. In healthcare, VAEs can assist patients with chronic conditions, providing them with emotional support, reminders, and motivational messages to promote adherence to treatment plans.

The development of VAEs has also led to the creation of new technologies, such as affective computing and social robotics. Affective computing involves the use of  machine learning algorithms to detect and analyze human emotions, enabling VAEs to respond accordingly. FAIR’s Seamless project aims to create emotionally accurate avatars with facial expressions, gestures, and body language. Social robotics, on the other hand, focuses on creating robots that can interact with humans in a socially intelligent manner, using ECAs as a key component.

Human face-to-face communication is a complex dance of verbal, vocal, and visual cues that require continuous adjustments. To model these dynamics, we are developing dyadic foundational models that capture the nuances of interpersonal interactions, including active listening, visual synchrony, and turn-taking. The Seamless Ineraction dataset, a large-scale collection of over $4,000$ hours of dyadic interactions, provides a valuable resource for training and evaluating these models. Building on this dataset, we develop a family of dyadic motion models that can generate facial and body motion, as well as respond to audio-visual inputs from users. These models showcase the potential for controllable and contextually relevant human-virtual agent interactions, paving the way for future research in social AI technologies, such as telepresence and multimodal video content analysis.

Virtual embodied agents have the potential to transform various aspects of our lives, from AI therapy and metaverse assistants to AI studio avatars and beyond. As research in this field continues to advance, we can expect to see more sophisticated and emotionally intelligent VAEs that enhance the communication between humans and machines.

\subsection{Wearable Agents}

 Wearable devices stand out from other smart devices in that they are equipped with cameras, microphones and other sensors on a device worn by the user to capture an egocentric perception of the physical world around the human user.  The unique nature of wearable devices distinguishes them from other smart devices, as they integrate AI systems that can perceive the physical world and help humans execute actions within it~\citep{10.1145/3637528.3672500}. This creates a synergy between perception and action, as wearable agents are embodied by the user, blurring the lines between human and machine.

The most notable application is AI agents in Meta's AI Glasses, see~(\url{https://www.meta.com/ai-glasses/}). Users can access Meta AI via these glasses to ask for information, access applications on the smartphone, and chat with an AI. Multimodal AI enables Meta glasses to see what the user sees and hear what the user hears in the environment. To realize the full potential of future wearable devices, various teams at Meta are working on wearable AI agents. AI agents in wearable devices must know how to plan actions in the physical world with reasoning. 

One approach is to prompt LLMs and VLMs for such planning. However, prompting LLMs or VLMs is limited in its effectiveness due to model inaccuracies and hallucinations.  As LLMs/VLMs are optimized to predict the next token or image pixels, they are inefficient in terms of long horizon action planning.  LLMs and Diffusion models have been shown to be ineffective according to a WordPrediction benchmark we devised. VLMs outperform both LLMs and DMs, but still hallucinate action plans. 

Meta researchers are working on different hypotheses, including an alternative world modeling based on transformer and JEPA architectures, as described in Section \ref{sec:wearable_architecture_models} below. This approach has the potential to enable more efficient and effective long horizon action planning, allowing wearable devices to provide more sophisticated assistance to users. Another approach is to train VLMs to directly predict a wearer's goals based on ego-centric context. To test this new set of models researchers created a benchmark of different scenarios that provide the needed contextual cues to perform goal inference. 
% TODO-FMe: Fix these citations/ references

Wearable agents have the potential to significantly enhance human performance by providing real-time guidance and support in various tasks. These agents can be categorized into two primary roles: coaching and tutoring. Coaching involves assisting users in physical activities, such as cooking, assembling furniture, or engaging in sports, while sharing the user's visual and auditory perspective. In contrast, tutoring focuses on mental world modeling, where an AI agent provides guidance and feedback to support cognitive tasks, such as mathematical problem-solving \citep{koedinger2006cognitive}.

To effectively coach or tutor, wearable agents require multimodal perception, enabling them to understand both the physical environment and user intent \citep{turk2014multimodal}. Moreover, these agents must be capable of responding to changing environments and user intentions in a non-deterministic manner, planning actions over extended periods \citep{Sutton1998}. This necessitates the development of AI models that can adapt to dynamic situations and exhibit machine initiative, rather than simply reacting to user prompts. While current large language models (LLMs) have demonstrated proficiency in solving mathematical problems, they are not designed to assist humans as tutors \citep{vaswani2017attention}. The development of AI tutors that can provide personalized guidance and feedback, without simply presenting solutions, remains an underexplored area of research. By addressing this gap, we can create more effective wearable agents that augment human capabilities and enhance learning outcomes.

The development of wearable devices has also led to new opportunities in fields such as education, healthcare, entertainment, and scientific research. For instance, wearable devices can be used to create personalized learning experiences, providing students with adaptive feedback and emotional support. In healthcare, wearable devices can assist patients with chronic conditions, providing them with emotional support, reminders, and motivational messages to promote adherence to treatment plans. In entertainment, wearable devices can enable more immersive and interactive experiences, allowing users to engage with virtual environments in a more intuitive and natural way. A wearable agent can assist experimental scientists in their laboratory work, and other scientists as research assistants. As research in this field continues to advance, we can expect to see more sophisticated and effective wearable agents that can provide personalized assistance, coaching, and tutoring to users.

\subsection{Robotic agents}

Robotic agents are AI systems used to operate a robot in the physical environment either performing tasks independently or through collaboration and interaction with humans. As opposed to specialized “single-function”-robots we review more general purpose robotic agents. The embodiment can take on various forms, however we focus on agents that can execute a diverse set of tasks in new environments. Such robotics agents are observing their environment, much like humans, via a variety of senses (for instance RGB cameras, tactile sensors, IMUs, force/torque sensors, audio sensors to name a few), and can control their robotic body via actions to achieve desired tasks. Their form factor varies from humanoid (bimanual with legs), to a more specialized form such as arms mounted on wheeled platforms. We envision that robotics agents have the potential to transform our world in at least two abstract ways: 1) They can learn to perform a variety of tasks and operate alongside us (often taking over dangerous or tiring tasks) or 2) they can form the foundation of learning general purpose intelligent agents via embodied interaction  in the real world.

Enabling robots to operate autonomously in unstructured environments collaborating with or supporting humans on daily activities is a long-standing dream of humans. Autonomous robots that are capable of acquiring general skills can help address societies in a variety of ways: Robots can help address labor shortages by taking over jobs that are physically or mentally demanding; they can be deployed in disaster scenarios to help with rescue missions (such as during natural disasters like earthquakes or tsunamis) - a job that is often dangerous for human rescuers; robots can support elderly care - which is often perceived as a more dignified way to help support the aging population; robots can support often overworked medical staff in hospitals [cite diligent ] and finally they can help free the regular people of household tasks to enable them to spend more time with family or on things they enjoy. To be successful, robotic agents will have to learn to autonomously perform tasks in environments built for humans, which fuels the investment in general purpose humanoid robots which mimic the human body (and with that the humans capabilities) as closely as possible.

While we often consider robotics agents primary purpose to be effective due to their potential to support humans in the real world through autonomous physical labor, the embodiment hypothesis posits that enabling robots to learn to interact with the real world is the only way to learn a general agent that can reason about the real world. The notion that machines can attain human-level intelligence through embodied learning in the physical world via sensory inputs, in addition to learning from records of human intelligence, remains an active area of scientific investigation. With the advent of large language models (LLM) as baselines, researchers can now investigate the benefits of embodied learning and its potential applications in the development of AGI. By integrating sensory inputs and real-world experiences into the learning process, researchers aim to create more sophisticated and human-like AI systems that can interact with their environment in a more meaningful way.

\section{World Models for Embodied Agents}
\label{sec:world_models}

World modeling is essential for embodied AI agents to understand and interact with their environment effectively. By creating a representation of the world, these agents can learn to reason, decide, adapt, and act, ultimately leading to more efficient, autonomous, and safe operation. World modeling can lead to higher efficiency, better task completion and enhanced safety. Embodied AI agents, such as robots or virtual characters, interact with their environment to achieve goals and complete tasks. To do so effectively, they need to understand the world around them, including its structure, dynamics, and relationships between objects. This is where world modeling comes in.

World modeling refers to the process of creating a representation of the environment that an embodied AI agent can use to reason about and make decisions. A physical "world model" needs to capture the relevant aspects of the environment, such as:

\begin{itemize}
    \item Objects and their properties (e.g., shape, size, color)
    \item Spatial relationships between objects (e.g., proximity, distance)
    \item Dynamics of the environment (e.g., movement, temporal changes)
    \item Causal relationships between actions and outcomes grounded in laws of physics
\end{itemize}

Meanwhile, an embodied agent needs to learn an internal representation of the human context via a mental world model. Such a model must capture the following aspects: 

\begin{itemize}
    \item Goals and intentions ( including their motivations, preferences, and values.)
    \item Emotions and affective states of the user and understanding how these emotions influence behavior.
    \item Capturing social dynamics, including relationships between individuals, groups, and institutions, as well as cultural norms, customs, and expectations.
    \item Understanding verbal and non-verbal communication, including language, tone, body language, and facial expressions.
\end{itemize}

By developing a mental world model that captures these aspects, an embodied AI agent can better understand human behavior, anticipate needs, and interact more effectively with humans in various contexts.

Embodied AI agents need world modeling in order to do:

\begin{itemize}
    \item Reasoning and planning: A world model allows an embodied AI agent to reason about the environment and make informed decisions. By understanding the relationships between objects and the consequences of its actions, the agent can plan and execute tasks more effectively.
    \item Zero-shot task completion: A world model enables an embodied AI agent to adapt to changing environments and tackle new tasks. By learning a representation of the world rather than rote-learning from text tokens and image pixels, an  agent can respond to new situations and unexpected events.
    \item Human-in-the-loop active learning: A world model provides a foundation for continuous and active  learning and improvement. By interacting with human users in the world, an embodied AI agent can refine its understanding of the world and improve its performance over time.
    \item Efficient Exploration: A world model helps an embodied AI agent to explore its environment efficiently. By focusing on areas of interest and avoiding unnecessary actions, the agent can gather information and learn more quickly.
\end{itemize}

World models enable embodied AI agents to reason and plan according to multimodal perception of the environment, user profile and preferences, previous actions and history. Hence they need perception models, physical and cognitive task planning, as well as contextual memory, as described in the following \S\ref{sec:Multimodal Perception} to \S\ref{sec:World Model Benchmarks}. 

\subsection{Multimodal Perception}
\label{sec:Multimodal Perception}

Multimodal perception and understanding from audio, speech, language, image and video is critical to embodied AI agents.  Agents take action according to what they perceive at each time step. 

\subsubsection{Image \& Video}
An embodied agent must have the capability of advanced image and video understanding. Such capability must be based on a  general-purpose vision-language models. 

At the core of image and video understanding is the Perception Encoder (PE), a state-of-the-art vision encoder trained using a simple yet powerful contrastive vision-language objective. Unlike traditional encoders that rely on task-specific pretraining, PE demonstrates that with careful scaling and robust video data integration, contrastive training alone can yield general visual embeddings that perform strongly across a wide range of tasks. PE models achieve state-of-the-art performance on tasks such as zero-shot classification, retrieval, question answering, and spatial reasoning. 

Building on this foundation, perception language models (PLMs) add the power of LLMs to build a general-purpose visual-language model. PLM, which leverages PE, is trained using large-scale synthetic and human-annotated image and video data. PLM model achieves comparable performance to state-of-the-art open-weight models without distilling from proprietary models. It outperforms fully open models and sets a new state-of-the-art in detailed visual understanding, with significant improvements in perception-focused image tasks (+9.1 points on average), video captioning (+39.8 CIDEr on average), and fine-grained video QA (+3.8 points on average). By combining transparent model design, high-quality data, and rigorous evaluation, PLM enables reproducible, state-of-the-art research in multimodal vision-language learning.

\subsubsection{Audio \& Speech}

Wearable embodied agents require sophisticated audio and speech understanding capabilities to effectively interact in their environments. Agents must be able to accurately detect and interpret ambient sounds~\citep{mesaros2024decadedcaseachievementspractices}, ongoing conversations~\citep{chime8} -- not directed to the agent -- and speech directed to the agent~\citep{yang2024mbestrqmultichannelspeechfoundation}. In many instances, multi-modal understanding can increase the effective signal-to-noise ratio or may even be required, e.g. to resolve the meaning of “this” in “what can I cook with this”. Critically, agents must also be able to address their environment with speech synthesis. Scaling deep learning-based speech processing to “Audio LLMs” is enabling wearable agents to take into account paralinguistics and better comprehend user intent and respond in real-time~\citep{défossez2024moshispeechtextfoundationmodel}. Any queries or actions that agents will support could be integrated into the audio LLM using “retrieval” and “tool use” capabilities. By leveraging these technologies, wearable embodied agents can provide more natural and intuitive user experiences in unrestricted domains and at high quality. Real-time capability, turn-taking behavior and the naturalness of the overall interaction (while maintaining factuality and other quality metrics) remain challenges for even the most advanced systems these days.

Most systems are based on a core LLM that has been pre-trained on large amounts of text. Input speech is being tokenized using an audio encoder that has been pre-trained on large amounts of audio data with the goal of well representing both semantic and para-linguistic context. Similarly, different schemes exist on the output side to faithfully synthesize words and increase the naturalness of the output given long-range context of the interaction.

\paragraph{Key Challenges}
\begin{itemize}
    \item Noise Robustness: Embodied agents are often deployed in noisy environments, making it essential to develop noise-robust speech recognition systems that can filter out background noise and focus on the user's voice. This may include multi-channel processing of audio, which can be used to improve the separation between primary speech and other ambient speech.
    \item Speech Variability: Users may speak in different accents, dialects, or languages, requiring wearable agents to be adaptable and able to understand diverse speech patterns. Separation of wearer's intended-for-service speech and speech meant for third parties (bystanders) is an especially hard problem, since it is all being emitted by the same source.
    \item Limited Computational Resources: Wearable devices often have limited processing power, memory, and battery life, making it crucial to develop efficient audio and speech processing algorithms.
    \item Tool Use, Retrieval Augmented Generation, and broad Factuality: while generative models have been shown to produce good-quality output on smaller domains, the extension to broad domains, while maintaining factuality and supporting tool use, has proven challenging.
\end{itemize}

\paragraph{Future Directions}
\begin{itemize}
    \item Edge AI: Developing wearable agents that can process audio and speech data locally, reducing latency and reliance on cloud connectivity. This also resolves concerns users may have around privacy.
    \item Personalization: Creating wearable agents that can learn and adapt to individual users' speech patterns and preferences. Such models will be able to better respond to user input and provide better experiences than general models.
    \item Multilingual Expansion: Enabling wearable agents to understand and respond to multiple languages, expanding their usability worldwide, possibly even to non-written languages or users that may not be able to read or write.
\end{itemize}

\subsubsection{Touch}

Humans (in fact all animals) have a sense of touch and heavily rely on it to purposefully interact with the environment and learn from their interactions. In humans,  touch provides an additional sensing modality that operates at a higher frequency than vision. It complements visual perception by providing a sense of contact. 

Especially when manipulating objects embodied agents will benefit from a sense of touch when occlusions are present (for instance when unpacking a bag, or when the hands itself occlude the object to be manipulated). However, the sense of touch not only gives a sense of contact, but also the force applied to the environment which is essential when manipulating objects that needs to be handled with care or when physically supporting humans (for instance elderly care). An important component of touch processing are general purpose touch encoders like \emph{Sparsh} \citep{higuera2024sparsh} that process touch signals and can enable force estimation, object slip detection, texture recognition, object pose and grasp stability prediction.

By addressing these challenges and leveraging these technologies, wearable embodied agents can provide more natural, intuitive, and effective user experiences.

\subsection{Physical World Models}

\begin{figure}
    \centering
    \includegraphics[width=0.4\linewidth]{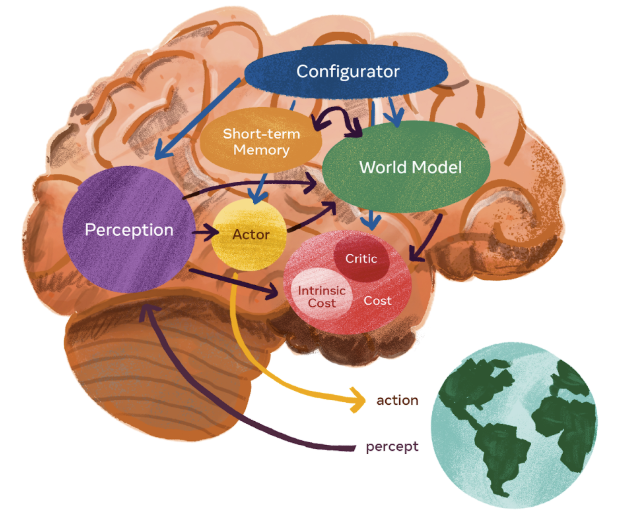}
    \caption{The modular system architecture for autonomous intelligence proposed by \citep{PathAMI}, illustrating its interaction with the world. Embodied AI agents introduce an additional dimension of interaction with the user.}
    \label{fig:autonomous_intelligence}
\end{figure}

To act in the physical world an embodied agent must understand this world. Planning is the practical reason we care about world models. An embodied agent that must achieve an explicit task  —  fix a bicycle, fold laundry, guide a user through a recipe—cannot rely on deterministic or memorized plans alone,  it must imagine how the environment will evolve under alternative actions (and changing environments and intent), score those imagined futures, and pick the sequence with the lowest expected cost. Supervised behaviour cloning can only imitate what it has seen; Reinforcement Learning (RL) requires trial-and-error in the real physical environment and explicit verifiable rewards which is extremely difficult to gather.  In comparison, LeCun’s AMI architecture \citep{PathAMI} lays out a clear operational paradigm: a \textbf{World Model} rolls the world forward under candidate actions, then a cost module evaluates each hypothetical future, and an MPC-style planner executes only the first step of the minimum-cost plan before re-planning with fresh observations.  A reusable world model lets an agent perform zero-shot planning in novel situations, refine its plan on the fly, and transfer across tasks without re-training.

Two design considerations shape every world model. The first is temporal and action semantic granularity. On the one hand we have low-level dynamics—joint torques that change every few milliseconds for robotic actions, while on the other hand we have human-scale actions such as “insert the battery”  that span seconds or minutes. The second consideration is the modeling approach. Generative models that  recreate the next observation in pixel space are expressive but expensive because they attempt to capture every low-level detail in the scene. LLMs with multimodal perception that are used to generate plans in text form are prone to hallucination due to spurious correlations in their training data. Conversely, joint-embedding world models forecast the world directly in an abstract latent space. They are efficient and stable. However the usefulness of joint-embedding world models depends on how well that abstraction captures the causal structure of the task.  

\subsubsection{Low-level motion planning}

\begin{figure}
    \centering
    \includegraphics[width=1\linewidth]{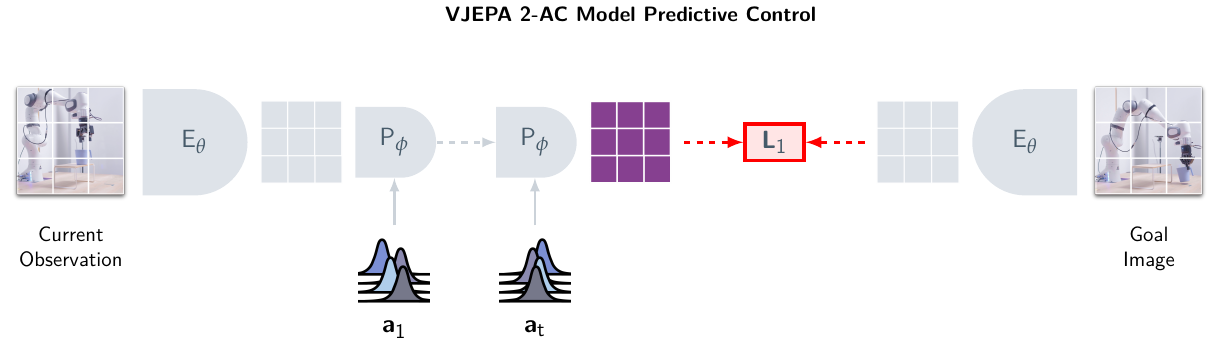}
    \caption{Planning with a JEPA visual world model for a robotic manipulation task. The future is imagined by encoding the context frames with the encoder $E_{\theta}$ and rolling out the predictor $P_{\phi}$ from these context visual embeddings and actions. To plan, any cost function with respect to the goal can be used. In this figure, the $L_1$ distance of the last imagined state to the goal state embedding is optimized with respect to the actions $(a_k)_{k \in [T]}$ using any trajectory optimization method, e.g. \citep{rubinstein1997optimization}. Taken from \citep{VJEPA2}.}
    \label{fig:vjepa2-mpc}
\end{figure}

Taking inspiration from how humans develop motion planning and control abilities, it seems sensible to use \textbf{learned world models} to send commands formulated at the level of the dynamics of the centroid of the body, of its limbs and of its manipulation tool (a hand or end-effector), below 20 Hz. Learned world models will a priori be too large to be called at a higher frequency. Such world models should send their commands to higher-frequency controllers, optimizing for the level of joint torques to satisfy the commanded actions, robot equivalent to muscle precise control, as discussed in \cref{subsubsec:classical-robotics}. This is the approach adopted in V-JEPA 2-AC~\citep{VJEPA2}, as illustrated in \cref{fig:vjepa2-mpc}, where the actions are formulated as a Delta in the end-effector’s (x, y, z) position and orientation. Then, the robot’s lower-level controller optimizes a trajectory to realize this commanded motion.

\paragraph{Visual world models.} Classical robotics mostly rely on proprioceptive information and analytical models to estimate the position of the robot and of objects in the environment.
A richer input to the robot’s controller is the visual input, which encompasses much more information than other sensor input used in classical control. Learning a world model from the visual input, as well as other potential proprioceptive or exteroceptive inputs, therefore seems indispensable to have a holistic and generic understanding of the physical world. Moreover, using a learned visual world model allows to make the connection with high-level world models, leveraging their semantic understanding of visual perception to plan more abstract tasks. As we will see afterwards, when using VLMs as high-level world models, one can plan in a space of abstract visual and textual representations that can guide the motion planner. The visual input is the highest bandwidth input, allowing babies to learn how the external world works and build their physical world model~\citep{IntPhys2019, bordes2025intphys, VEO}.

\paragraph{Planning with world models.} In planning, we can use a learned predictive model to plan towards a goal state, or to minimize a cost function on the future planned states. The cost function can include a term that is a distance to a goal state (in \cref{fig:vjepa2-mpc}, we take the distance between embeddings) as well as other penalties and objectives to minimize. The optimizer over the future actions can be gradient-based, which is possible when the learned model is differentiable or non-gradient based, as in the commonly used method from \cite{rubinstein1997optimization}.
% (see Basile’s JEPA-WMs, and ongoing work by Amir Bar and Michael Psenka) 
Planning (system 2) is more costly than calling a policy (system 1) but is more likely to generalize zero-shot outside the training distribution. Learned world models have been used for both use cases: 1) training a policy on trajectories imagined with the world model, 2) using the learned model for model-based planning.

\paragraph{Generative world models.}To model the dynamics of the physical world from the visual input, generative video models are trained to generate the most likely future frames, given no conditioning \citep{Magvit, kalchbrenner2016videopixelnetworks,babaeizadeh2018stochastic,denton2018stochasticvideogen} or conditioning on a text prompt ~\citep{Gupta2023PhotorealisticVG,villegas2023phenaki,Kondratyuk2024VideoPoetAL} or on actions ~\citep{oh2015actioncondvideopred,chiappa2017recurrentenvsims, bruce2024genie, parkerholder2024genie2}. Although such models have been used to train policies in simulated videos~\citep{UniSim, DreamerV3}, the potential of these models to control agents in a zero-shot manner is not clear yet.

\paragraph{Joint-Embedding predictive world models.}Using pretrained visual encoders and training a world model to predict in this embedding space is the most widespread approach to learn visual world models. Video generation models typically rely on continuous or discrete VAEs (VAE, VQ-VAE) to achieve high visual fidelity. In contrast, world models intended for downstream tasks such as controlling agents or predicting future segmentation maps benefit from predicting the future in a high-level latent space ~\citep{PathAMI, VJEPA2, DINO-Foresight, DINO-WM, PLDM}.

\subsubsection{High-level action planning}

High-level action planning refers to the ability of an embodied agent to generate, organize, and reason over sequences of actions that extend over long temporal horizons which actions that have a higher semantic abstraction. If low-level models enable an agent to grasp an object or navigate a short trajectory, high-level models allow the agent to achieve complex, goal-directed tasks such as preparing a meal, assembling a device, or guiding a human user through multi-step procedures.

These models operate at human-relevant temporal scales—often on the order of seconds or minutes—and must account for causal dependencies, temporal ordering, and task decomposition. Instead of reacting reflexively to immediate inputs, agents with high-level planning capabilities anticipate future states, infer intentions, and coordinate action sequences that unfold in contextually meaningful ways.

A key affordance of high-level planning is abstraction. Rather than reasoning over raw sensory data or low-level motor commands, high-level planners work over symbolic, linguistic, or otherwise abstract representations that compactly encode semantic state transitions. For instance, rather than understanding every motion-level transition, a planner might reason about transitions such as “the drawer is open” or “the battery is installed.” This abstraction dramatically reduces the size of the planning search space and improves generalization across tasks, as abstract actions tend to recur across different contexts.

This level of reasoning is essential in embodied AI applications ranging from assistive robotics to augmented reality. However, several challenges still remain in high level action modeling. First, the diversity and richness of procedural activities make high-level actions hard to simulate or to enumerate task variations. Any synthetic environment only allows a certain set of actions which is task, domain and environment specific. Due to this, the integration with physical context is another challenge, especially in dynamic, real-world environments. While a motion-level or manipulation-based action has a clear consequence that is easy to perceive, a high-level action could induce consequences on the environment that are hard to dynamically simulate. Lastly, evaluation of high-level action planning does not have an objective measure of procedural plausibility or task success in open-ended environments. Even in closed environments, metrics such as success rate, mean intersection over union or mean accuracy over fixed and short (three to four steps) horizons show very little generalizability to real world use cases. As such, high-level action planning lies at the intersection of perception, reasoning, and language understanding, and requires models that can simulate procedural knowledge on the physical world.

\begin{figure}
    \centering
    \includegraphics[width=0.95\linewidth]{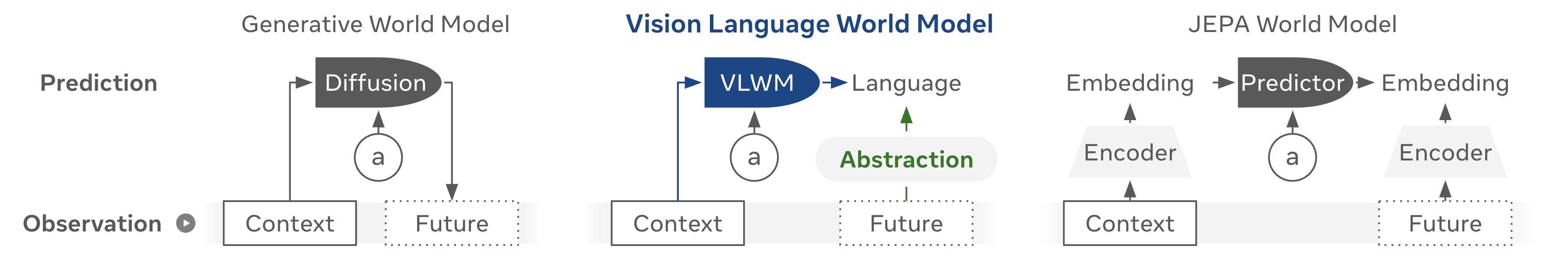}
    \caption{Comparison of action ($a$) conditioned world model architectures. VLWM is a JEPA-style world model that predict abstract representation of future world states, instead of generating noisy and high-volume raw observations.}
    \label{fig:enter-label}
\end{figure}

\subsection{Mental World Models}
Mental world modeling in humans  is the process of creating a mental representation of the world, including objects, events, and relationships \citep{johnson1983mental}. This cognitive ability plays a crucial role in reasoning, as it enables humans to simulate scenarios, predict outcomes, make counterfactual and casual reasoning, in order to make informed decisions. They are also a form of abstraction and generalization. Humans learn to mentally model the world through evolution and various forms of embodied, experiential and supervised  learning. We propose that it is necessary for agents to learn the mental states of humans in order to assist and collaborate with us better. 

Whereas physical world models are internal representations that AI agents construct to understand, predict, and reason about the external world, a mental world model is the AI agents’ representation of mental states of human users or other AI agents, which is essential for Theory of Mind (ToM) reasoning. Mental world models are particularly important for human-agent interactive tasks, such as assistance or tutoring, as well as multi-agent collaborations.

A mental world model typically consists of several key components, including:

\begin{itemize}
    \item \textbf{Beliefs}, which are representations of human knowledge or opinions about the world

    \item \textbf{Goals}, which are representations of the human user’s  desired outcomes or objectives

    \item \textbf{Intentions}, which are representations of human or agent’s plans or actions to achieve their goals

    \item \textbf{Emotions}, which are representations of the human’s emotional state, influencing their behavior and decision-making

\end{itemize}

By incorporating these components, mental world models can be applied in various ways, such as:

\begin{itemize}
    \item \textbf{Agents can anticipate goals and intentions}. By anticipating a user's goals and intentions, a mental world model can enable the agent to proactively offer assistance or guidance to help the user achieve their objectives.
    \item \textbf{Agents can infer belief discrepancies}. Given a dialog where one person mistakenly believes an object is in a certain location while another knows the truth, a mental world model can infer the belief discrepancy and predict how the person with false-belief will act.
    \item \textbf{Agents can predict emotional responses}. A mental world model can predict how a user will emotionally respond to a particular message or action, allowing the agent to adjust its approach to better support the user's needs.

\end{itemize}

The benefits of a mental world model imbued into AI agents are many. They include:
\begin{itemize}
    \item Mental world models \textbf{facilitate collaboration} between humans and AI systems by enabling agents to represent and reason about users beliefs, goals, values, and preferences, leading to more efficient and effective human-AI collaboration.

    \item Mental world models can enable agents to better understand user mental states. This can lead to more \textbf{effective and empathetic interactions}.

    \item Mental world models allow agents to \textbf{proactively and strategically plan} and select their conversational moves, guiding users towards their goals and improving task effectiveness.
\end{itemize}

Given a perceptual stimulus input $X_p$ with (one or more) human subject(s) (e.g. images, videos), the mental state $X_{m,s}$ of a subject s in X consists of textual descriptions of his/her:

\begin{itemize}
    \item Attention to certain perceptual fields
    \item Short-term memory
    \item Long-term memory (including relevant world knowledge)
    \item Short-term intentions
    \item Long-term goals
    \item Emotions
    \item Beliefs about the physical world state
    \item Beliefs about mental states of the other subjects
\end{itemize}

This is not an exhaustive list, and the elements/dimensions of the mental state modality can be flexibly adapted to application scenarios. 

\begin{figure}
    \centering
    \includegraphics[width=1\linewidth]{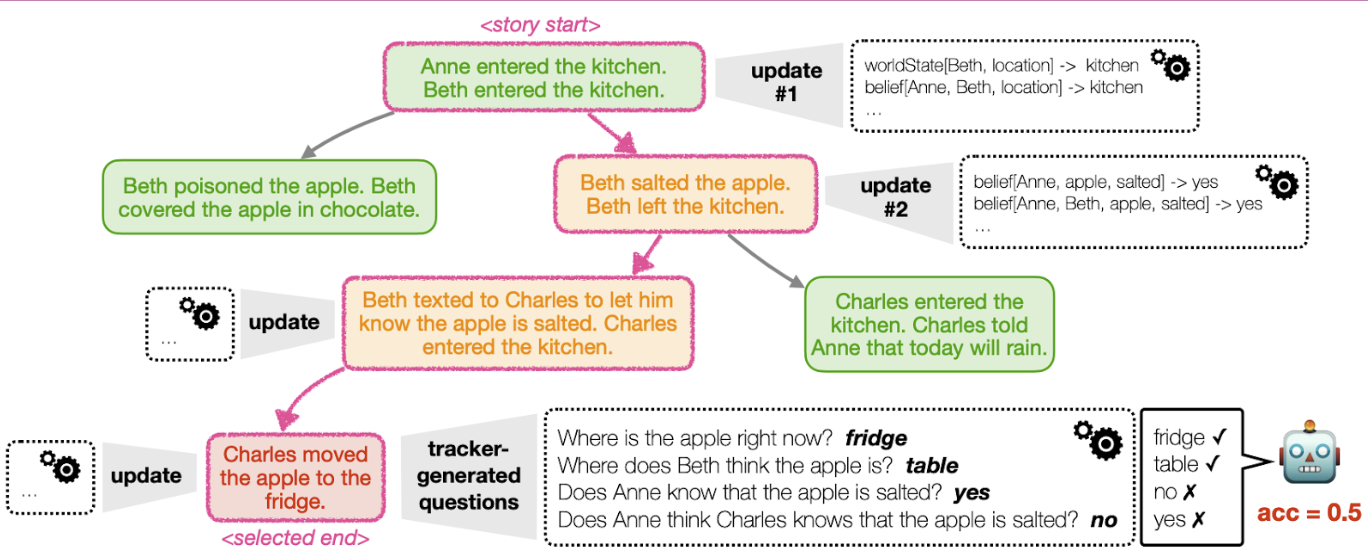}
    \caption{Probing nested beliefs through program guided scenarios. We adversarially generate belief reasoning tasks via ExploreTom~\citep{exploretom2024}.}
    \label{fig:enter-label}
\end{figure}

Several recent works have highlighted the limitation of current AI models (such as large language models) in understanding and reasoning about others’ mental states. Benchmarks like ToMI~\citep{Le2019}, and Hi-ToM~\citep{wu2023} focus primarily on mental state tracking, but have significant limitations: ToMi only supports a restricted set of actions, while Hi-ToM extends it slightly, both datasets have extremely restricted interactions to orders.  Recent work on ExploreToM~\citep{exploretom2024} advances this line by adversarially generating belief-based reasoning tasks through programmatic scene construction. By modeling theory of mind as an active probing challenge, where agents must ask questions or interpret belief updates, ExploreToM stresses the need for structured representations that evolve over the course of interactions. This direction complements our broader agenda on mental world models, which moves beyond single-turn inference toward agents that build and refine internal models of others during extended collaboration tasks. Even though ExploreToM focus on evaluation, future works should investigate works that integrate latent belief tracking, and social feedback into the learning loop, aiming to provide agents capable of dynamically adapting to others’ perspectives in open-ended settings.

\subsection{Actions and Control}

Different embodied agent types carry out actions in different spaces:  2D virtual agents in the digital space. 3D virtual agents carry out actions in the augmented reality (AR) or virtual reality (VR) environments.  Wearable agents plan procedurally in the real world and show or tell a user what to do and when. Robotic agents must control robotic actions in the physical world. 

\emph{Virtual embodied agents} are rendered in digital environments and vizualised using VR headsets or AR glasses to have personal interaction with users. Such agent typically use controllable motion models that can adapt emotional responses and expressitivity levels for face features, and generate gestures for body language. Social intelligence is an important capability of VEAs, for which the agent has to control the interplay of verbal (speech model) and non verbal (motion model) actions.

\emph{Wearable agents} advise human actions by “showing” in visuals and “telling” in speech, or carry out actions on smart devices. Current human-agent interaction paradigm is user-initiative where the human issues a task prompt and the agent carries out actions end-to-end. Some studies also show the benefit for agents to react to the context and carry out actions autonomously in a machine-initiated manner. We propose that, in advising human actions,  an agent needs to know not just what to show and what to tell, but when to show and when to tell, in a fully mixed-initiative manner. Embodied agents need to be fully interactive with humans, allowing direct queries as well as predicting when the user needs guidance. 

\emph{Robotic agents} typically control the robotic hardware either directly at the joint level (sending position, velocities and/or forces/torques to each actuator of the robot), or at a more abstract level such as desired (relative) position of the robots trunk/base, the (relative) position of the end-effectors (hands), and the (relative) position of the feet (if applicable), which are then translated to joint level control commands. For each action space, the robotic system has controllers implemented that can compute the low-level command (often desired joint forces/torques) from the chosen action space at the needed frequency for safe robot operation.

\subsection{Memory}

One core ingredient of the world models involved in Embodied agents is Memory. It is involved in the processing of the agent, and in particular it captures the interaction of the agent with the world, and consolidates it into an internal representation, which effectively parametrizes a world model. Here, we briefly review the kind of memory included in the transformer architecture and the key features and challenges inherent to producing a memory that could empower an embodied AI agent.  
The position paper by~\cite{pink2025position} advocates the importance of episodic memories for agents. It provides a complementary view on different types of memories.

\subsubsection{Existing memories}

In popular neural networks, we can distinguish 3 forms of memories:

\paragraph{Fixed memory: The model weights.} This form of memory is defined by the parameterization of the function corresponding to the neural network model. They are learned by training on a pre-defined (fixed) dataset, and as such we expect that they incorporate the knowledge appearing in it. Once training is complete, these weights are fixed at inference time. This comes with two limitations: 
\begin{itemize}
    \item In order to add some new knowledge corresponding to additional training data, or to adapt to a new task or a new domain, one can do some fine-tuning. This typically comes as a trade-off between acquiring new knowledge and not forgetting the knowledge injected during the pretraining stage. In practice, for conversational agents, fine-tuning is an effective way to perform alignment, i.e., adapting the model so that it is better aligned with what we expect from the model, including making the model safer.
    \item The number of weights is fixed beforehand, which means that the capacity is constant. Fine-tuning or LoRA do not add capacity to the model.
\end{itemize}

\paragraph{Working memory.} These working memories are typically a subset of what we usually refer to as activations. Mathematically, viewing a neural network as a composed function, they are the output and input of intermediate functions. As such, the writing process amounts to computing these intermediate functions. This is fast, in the sense that it does not require multiple passes on the data like with back-propagation. Examples of such working memories typically include specific subsets of activations: the internal state of Recurrent Neural Network, State space Models (SSM). For transformers, this corresponds to the KV-cache. 

In this context, we distinguish between mutable and immutable memories. A mutable memory, for instance the internal state of a RNN, can be updated and modified. An immutable memory is not modified once calculated, which makes it possible to index and cache, like the KV cache in a transformer for a given timestamp. The write and read procedures are also simple, since there is no need to address the balance between adding new content to the memory and not forgetting. The downside is that it is not compressed and therefore needs to grow linearly (with a large constant) to keep track of all potentially useful information, while also requiring an increasing amount of Flops for the read operation. 

\paragraph{External memory.} This last kind of memory consists of brute information stored out of the architecture, and accessed with a specific mechanism, in particular it includes Retrieval Augmentation Generation (RAG), which is probably the most prominent form of external memory. In some ways, chains of thoughts can be regarded as a form of external working memory. In this case, the input and output space themselves serve as the working memory.
External memory is typically uncompressed, yet typically accessed  through a search with embedding that offers fast access to a subset of relevant subset of the dataset. 

\subsubsection{Challenge and objective of Memory for world models}

The research in memory requires creating a new form of memory, that we refer to ‘’episodic memory’’, that could grow in a scalable way when interacting with the environment. 

\paragraph{Limitations of current memories.} The aforementioned set of memories all come with their limitations. Notably the fixed memory is bounded and has a slow writing operator (back-propagation). In contrast, KV-cache grows linearly with the time of interaction with the user, which does not scale in time as it will eventually become too bulky. 
The external memory is currently one of the simplest way to keep track of the interaction with a user without exploding the computational complexity. However it remains uncompressed, hence requires to store all interactions in a database. Additionally it comes with additional limitations, in particular it requires some intermediate processing to leverage the store knowledge. 

An episodic memory would answer to multiple goals, which we describe below.

\paragraph{Personalization.} Augmenting an existing architecture with an explicit memory is a simple way to offer personalization, since it is possible to dedicate specific zones of the model to personalization. This problem is akin to model adaptation and to solution developed in that respect, including adapters~\citep{rebuffi2017learning,signe2024neutral} or LoRA adapters~\citep{hu2022lora} or LoRA. One aspect that is especially important for agents is how to make personalization more resource efficient, both from a memory and training point of view. Ideally, personalization would maintain a subset of values specific to the user allowing for storing the history of the interaction in a compressed manner.

\paragraph{Life-long training.} Existing architectures and training procedures heavily rely on the pre-training / post-training / inference paradigm: during pretraining, a proxy objective is employed to incorporate vast amounts of knowledge in the model. The post-training stage(s) makes the adaptation to the use-case and ensures that the model better aligns with the use-cases. At inference time, the model is frozen and only the kv-cache can be customized. 
Our research will help to go beyond this segmentation, and in particular will ensure that the model can learn forever once it starts interacting with its environment and users. Currently this is not possible because the resources required for that are growing linearly with interaction time.   

Hence, one challenge of long-life training is that it requires a mechanism to grow the memory capacity in a sublinear manner. We regard this design choice as a necessity to store \textbf{World Models}. In that respect, we will investigate test-time training strategies, as proposed in multiple recent works on memories, as well as forward-read-write methods that could leverage local update rules. 

\subsection{World Model Benchmarks}
\label{sec:World Model Benchmarks}

\subsubsection{Minimal Video Pairs}
The Minimal Video Pairs (MVP) \citep{krojer2025shortcut} benchmark is designed to provide a more robust assessment of video-language models’ physical and spatio-temporal reasoning capabilities by addressing a common pitfall in existing evaluations—performance inflation due to shortcut exploitation. MVP consists of 55,000 multiple-choice video question-answer pairs that focus on understanding physical events across a range of real-world and simulated contexts, including egocentric and exocentric video, robotic interaction, and cognitive science-inspired intuitive physics tasks. A key feature of the benchmark is the inclusion of minimal-change video pairs: each pair consists of two nearly identical videos paired with the same question but requiring different correct answers. This design forces models to rely on fine-grained physical understanding rather than superficial visual or linguistic patterns. The benchmark reveals a significant performance gap between current video-language models and humans—state-of-the-art models achieve only 40.2\% accuracy, well below the 92.9\% human baseline and close to the 25\% random guess rate—highlighting persistent challenges in physical reasoning and generalization.

\begin{figure}[h!]
    \centering
    \includegraphics[width=0.75\linewidth]{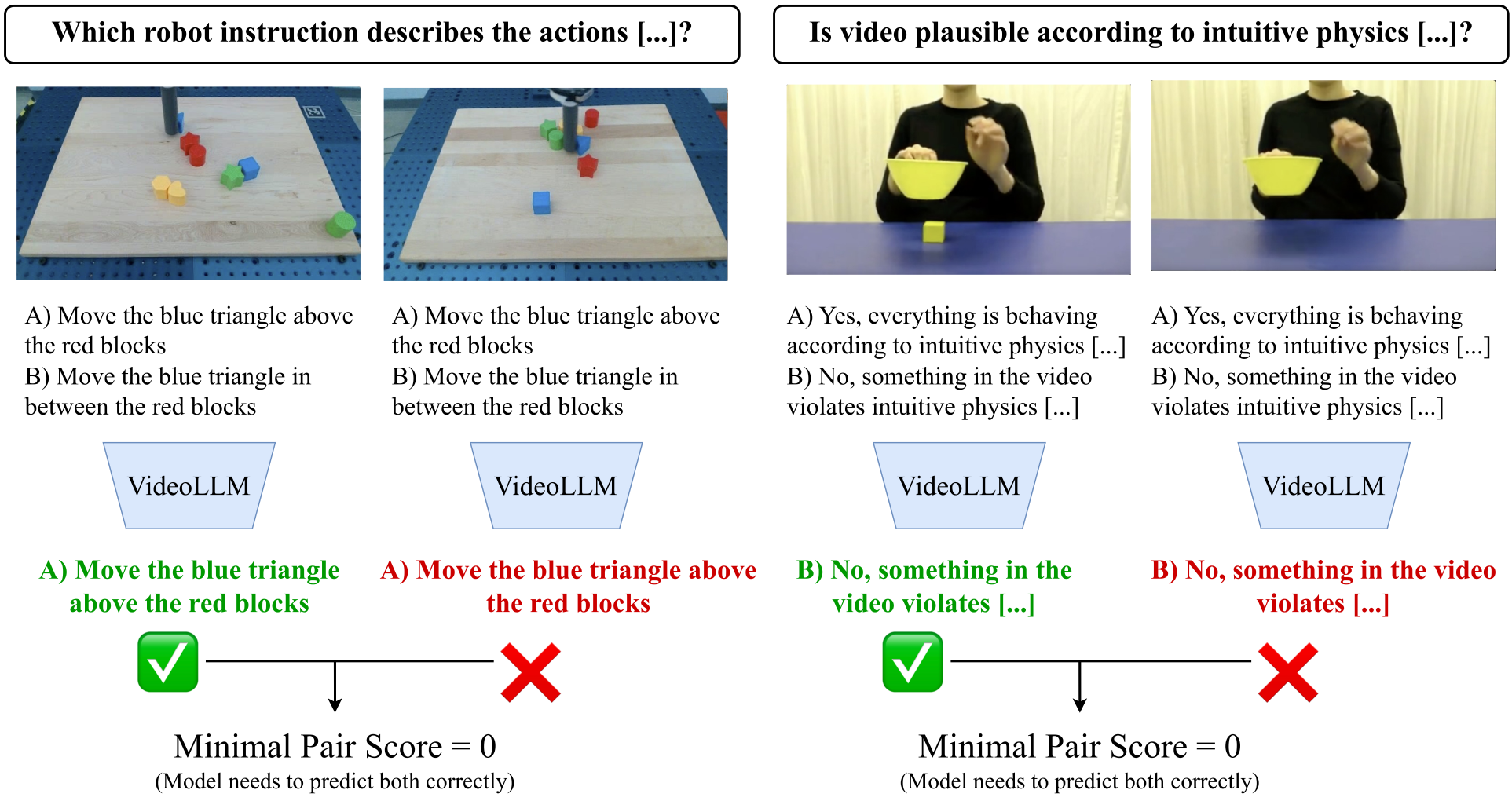}
    \caption{Minimal Video Pairs Benchmark (MVPBench) evaluates video-language models on their ability to understand and reason about the physical world using questions paired with two visually similar videos with differing answers.
    }
    \label{fig:enter-label}
\end{figure}

\subsubsection{IntPhys2}
IntPhys 2 \citep{bordes2025intphys} is a video-based benchmark designed to assess a model’s grasp of intuitive physics, building on the earlier IntPhys framework. It targets four fundamental principles governing macroscopic objects —Permanence, Immutability, Spatio-Temporal Continuity, and Solidity— drawing inspiration from cognitive developmental studies of early childhood reasoning. Using a violation-of-expectation paradigm, IntPhys 2 presents controlled virtual scenarios that contrast physically plausible and implausible events, challenging models to recognize and reason about these discrepancies. Benchmark results show that while contemporary vision models can process basic visual features, they struggle with deeper physical reasoning, often performing at chance level across complex test cases. In contrast, human participants approach ceiling performance. These findings highlight the limitations of current systems in capturing core physical intuitions and underscore the importance of more sophisticated modeling approaches for achieving human-like understanding in embodied agents.

\begin{figure}[h!]
    \centering
    \includegraphics[width=0.6\linewidth]{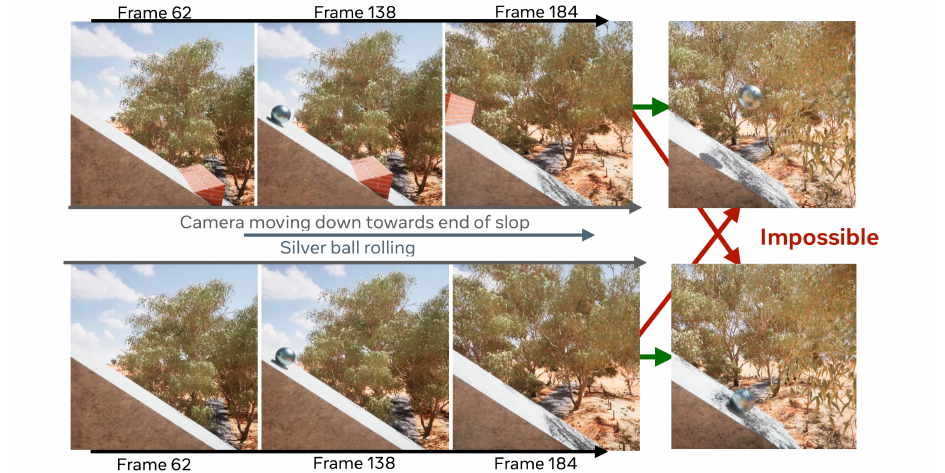}
    \caption{IntPhys 2 benchmarks a model's understanding of intuitive physics using a violation-of-expectation framework.
    }
    \label{fig:enter-label}
\end{figure}

\subsubsection{CausalVQA}

CausalVQA \citep{foss2025causalvqa} is a benchmark aimed at evaluating video question answering (VQA) systems through the lens of causal reasoning in real-world contexts. Unlike prior VQA benchmarks that either emphasize low-level perceptual understanding or focus on synthetic environments for physical reasoning, CausalVQA bridges this gap by presenting diverse and challenging questions rooted in natural video scenarios. The benchmark includes five distinct question types—counterfactual, hypothetical, anticipation, planning, and descriptive—that collectively assess a model’s ability to reason about causal relationships, predict outcomes of actions, and understand physical dynamics over time. To ensure robustness, the dataset incorporates design strategies that discourage shortcut exploitation, thereby requiring models to rely on genuine visual and temporal comprehension rather than superficial textual patterns. Empirical evaluations show that state-of-the-art multimodal models significantly underperform compared to humans, particularly on tasks involving anticipation and hypothetical reasoning, underscoring the limitations of current systems in capturing causal and temporal structure, which are core capabilities for effective world modeling in embodied agents.

\begin{figure}[h!]
    \centering
    \includegraphics[width=1\linewidth]{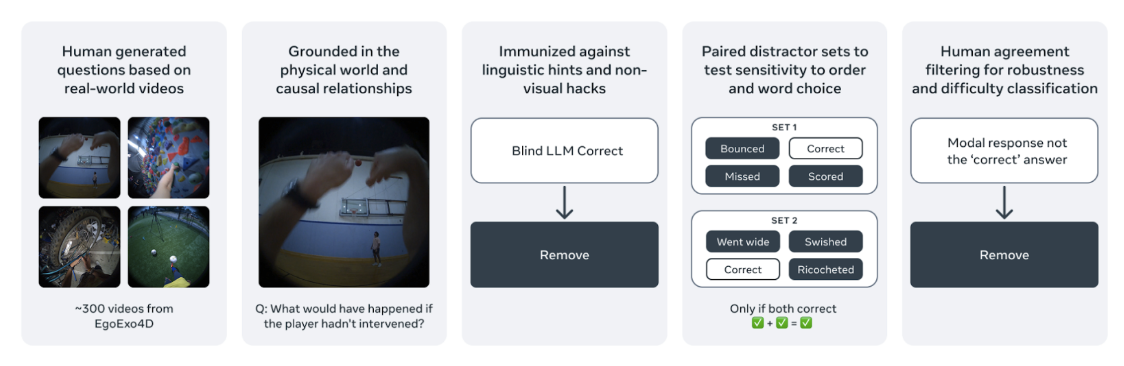}
    \caption{CausalVQA Benchmark evaluates models' understanding of causality through QA pairs grounded in real world videos.}
    \label{fig:causalvqa}
\end{figure}

\subsubsection{World Prediction Benchmark}
\label{sec:world_prediction}

Evaluating the capacity of AI models to perform high-level action planning requires benchmarks that reflect the kind of world modeling humans naturally perform—inferring and reasoning about abstract world states to guide behavior. While prior benchmarks in AI have largely concentrated on low-level world dynamics or robotic control, the WorldPrediction \citep{chen2025worldpredictionbenchmarkhighlevelworld} benchmark introduces a new evaluation framework centered on procedural planning and temporal abstraction. This is particularly relevant for wearable agents, which must interpret dynamic visual environments and anticipate user needs in order to provide timely and contextually appropriate assistance. WorldPrediction comprises two tasks: identifying the correct action between initial and final states (WorldPrediction-WM) and selecting the correct sequence of actions from distractors (WorldPrediction-PP). Unlike datasets that rely on low-level visual continuity, this benchmark incorporates “action equivalents” to isolate high-level reasoning from perceptual shortcuts. Built on a formal partially observable semi-MDP framework, it supports robust comparisons across AI systems, including generative models. Extensive human validation confirms the benchmark’s difficulty: while humans perform near perfectly, current state-of-the-art models reach only 57\% accuracy on WM and 38\% on PP.

\begin{figure}[h!]
    \centering
    \includegraphics[width=1\linewidth, trim=0 100 0 0, clip]{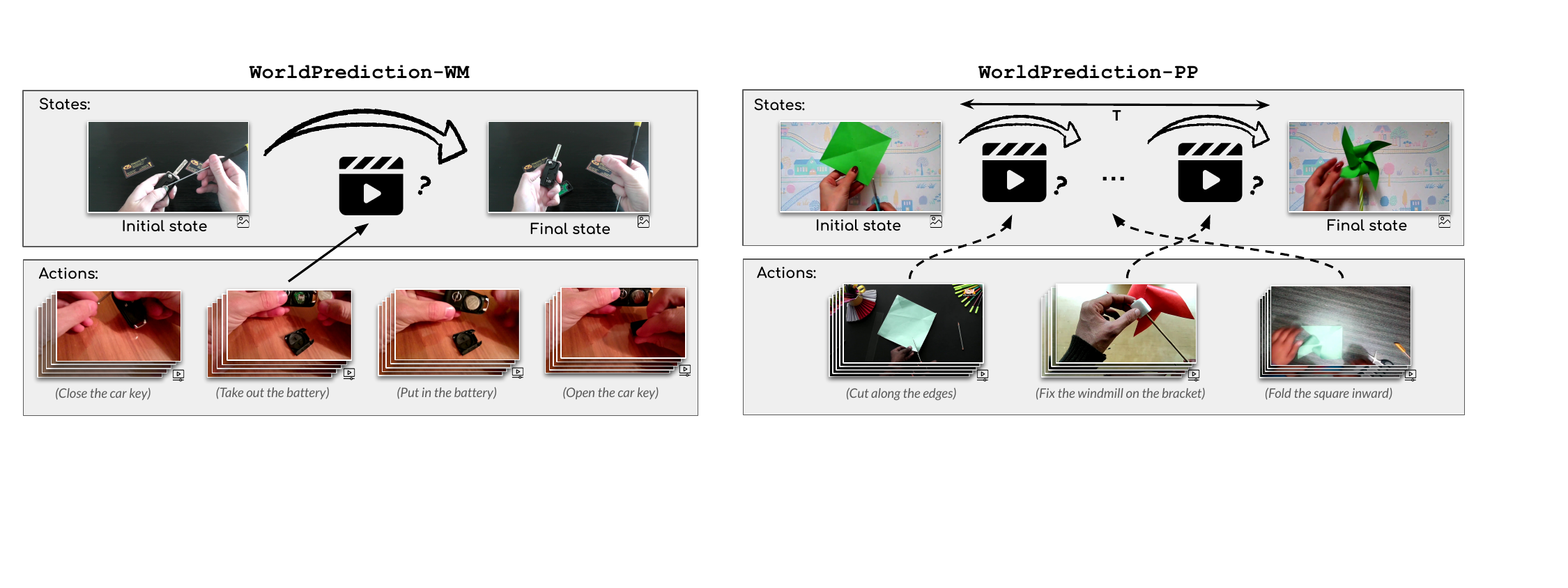}
    \caption{WorldPrediction benchmark evaluates the high-level world modeling and long-horizon procedural planning capabilities of models from observations only, where the actions are depicted as videos and the states as images.}
    \label{fig:worldprediction}
\end{figure}

\section{Type I: Virtual Embodied Agents }
\label{sec:type_1_virtual_embodied_agents}
Conversational AI has an illustrious history of assisting human tasks. Its development gave rise to many generations of AI assistants, from call-centre agents, to virtual assistants on smart devices such as Siri, Alexa, etc. In the 2010s, thanks to the increasing power of neural models, chatbots were developed to purely carry out conversations with human users in multiple turns. \citep{shuster2022blenderbot}. As the underlying encoder-decoder architecture keeps improving, especially following the transformer architecture, there was a step change in the quality of these conversational agents. ChatGPT was the first LLM fine-tuned on conversational data, that is not only able to carry out conversations with users but also showed emergent multitasking capabilities \citep{bang2023multitask}. Conversational AI agents that are targeted for various tasks have emerged in the last year as “AI agents”. However, like the  Siris and Alexas, they are  not embodied in any physical form.

Meanwhile, we can recall Microsoft Tay and Xiaobing, both were “personified” in the character of a teenage girl, and with a cartoon human face. This type of conversational AI agents embodied in virtual characters have been developed, in parallel to the class of “faceless” conversational AI agents, to enhance human-agent interactions.  Studies \citep{liu2023human} have shown that humans relate better to avatars that can emote, than to faceless conversational agents such as chatbots, when the capabilities are equivalent. 

AI avatars, unlike the cartoon faces of yore, are designed and programmed to be capable of emoting with proper facial expressions and speaking with proper lip formations, to mimic that of a human agent. Their face can be modeled using code controls \citep{cao2019openpose,ekman1978facial}, with deformation driven by learned of hand-crafted face \citep{SMPLX}.
Similarly lifelike body appearance (body, hair, clothes) and gesture can be captured or learned from visual inputs \citep{baran2007automatic,matsuyama20123d,lombardi2018deep, huang2020arch,armando20234dhumanoutfit}

Prior study \citep{ma2019exploring} has shown that one can improve an avatar’s emotional intelligence by incorporating personality-driven emotional expression. In other words, endowing with a consistent, distinctive personality trait conveys a stronger sense of emotional intelligence and mitigates user challenges better.  Another study \citep{liu2023human} has shown that AI avatars, a new phenomena in AI agents,  offer advantages such as enhanced trustworthiness and increased adoption, enabling human-like interaction and engagement. However, they also raise concerns about psychological impact, discrimination, and biases.

Another important dimension to consider is that AI embodied avatars could interact with the users in VR and MR environments. Imagine a virtual companion that can play games in VR or being a virtual pet in mixed reality. In this case, AI avatars need to be capable of perceiving and understanding their surrounding environment and interacting with it in a way which is coherent with their embodiment.

One approach to train virtual embodied agents to effectively interact with their environment is to use the reinforcement learning (RL) paradigm, which focuses on learning “policies” to control the body of the agent to optimize a reward function that formalizes a desired task (e.g., win a game). This approach is popular in physics-based character animation, where a physics simulator is paired with a RL algorithm to generate fully reactive behaviors with a high-level of realism. This holds the promise of fully automatizing the creation of AI avatars capable of realistic and consistent behaviors and solving complex tasks.

More recently, Seamless Interaction \citep{seamless_interaction} introduces a family of audiovisual dyadic motion models compatible with both 2D and 3D renderings. The models, conditioned on speech from two parties as the input, can jointly generate facial expressions and body gestures. The motion models produce gestures and expressions of one specific speaker while taking into consideration the audio from both people. This allows the models to visualize speaking gestures, listening gestures, and turn-taking cues. The AV Dyadic Motion models go one step further to show visual synchrony by also taking into consideration the visual input of the other party.

\subsection{Capability Taxonomy}

The virtual agent boasts a range of cutting-edge capabilities that enable it to deliver highly realistic and engaging interactions. One of its key capabilities to generate full-body movements, including facial expressions, arm and hand gestures. This allows the agent to convey emotions and intentions in a more nuanced and human-like way.
Another significant feature of a virtual agent is its dyadic audiovisual conditioning capability. This means that the model can take into account speech from two parties, whether it is human-generated or produced by a speech large language model (aka, speech LLM). Additionally, the agent can optionally condition on visual input from users, enabling interesting visual synchrony phenomena such as smile mirroring. This feature enables the agent to respond more naturally and empathetically to user input.
The virtual agent also offers face controllability, allowing models to be fine-tuned for greater expressiveness. This has significant potential applications in building more attentive and empathetic virtual listeners, which can be particularly valuable in fields like customer service and education.
In terms of gesture generation, virtual agents could produce illustrative gestures that match the semantic content of the conversation. For example, when saying the word "fly," the agent will extend its arms to mimic the action of flying. This adds an extra layer of realism and engagement to the interaction.
Besides motion generation, the research of virtual agents has made significant strides in integrating motion models with speech LLM which play a role of spoken dialogue generator. The output of a speech LLM can visually guide the motion of the agent, creating a more seamless and natural interaction. This integration of speech and motion models paves the way for even more sophisticated and intuitive interactions in the future.
Finally, the output representations of motion models should be visualized by the virtual agent. For example, virtual agents could use 2D and 3D renderings to demonstrate visualizations, which is critical for a fully immersive VR experience.

\subsection{Architecture and Models}
\begin{figure}[h]
    \centering
    \begin{subfigure}[b]{0.8\textwidth}
    \centering
    \includegraphics[width=0.8\linewidth]{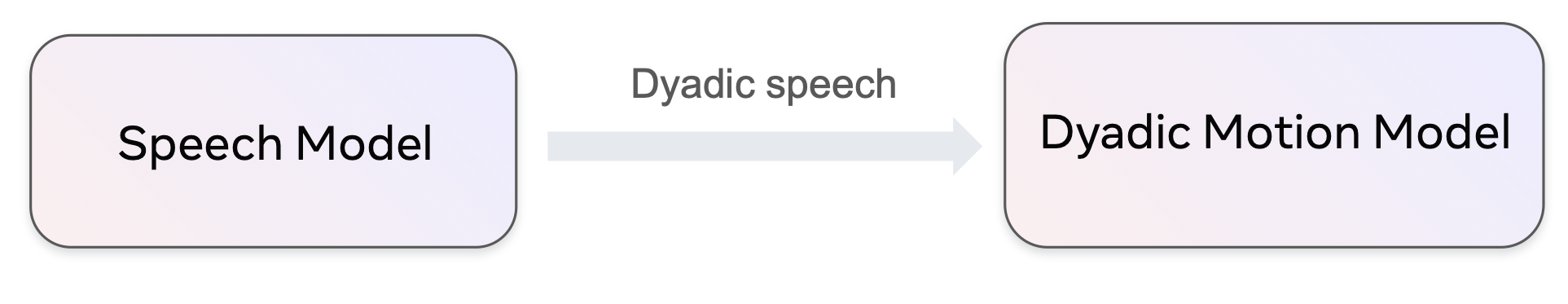}
    \caption{Cascaded integration.}
    \label{fig:speech_lm_motion}
    \end{subfigure}
    \begin{subfigure}[b]{0.9\textwidth}
    \centering
    \includegraphics[width=1.0\linewidth]{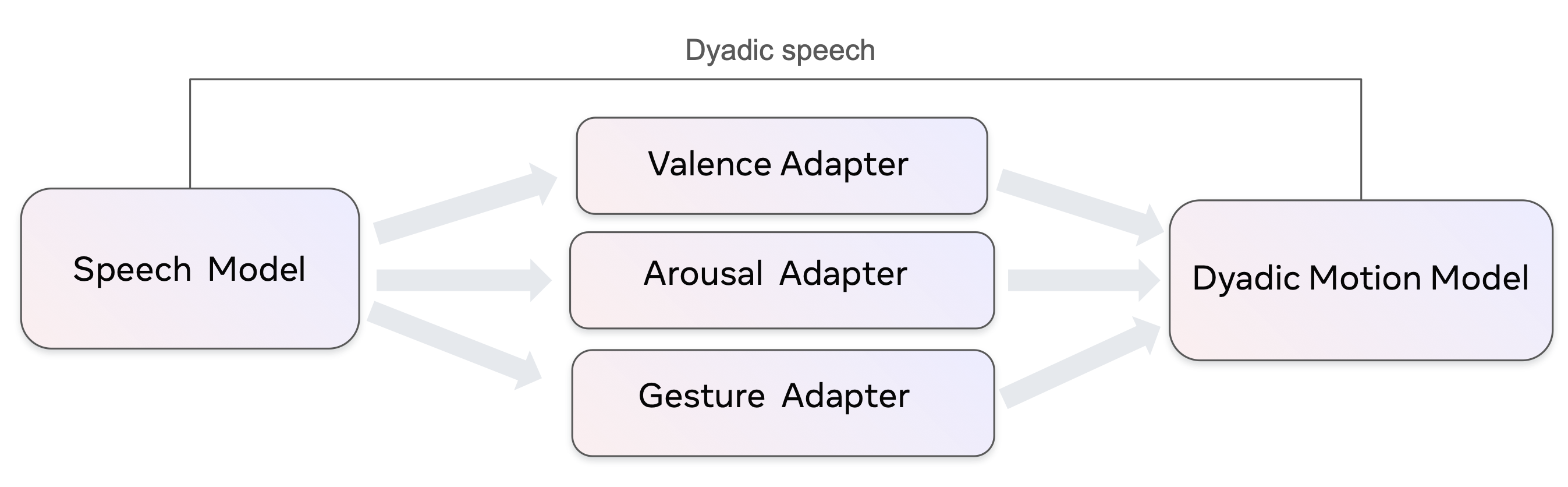}
    \caption{Codebook integration.}
    \label{fig:speech_lm_code_motion}
    \end{subfigure}
    \caption{Architecture of our Dyadic Motion Model.}
    \label{fig:speeech_lm_motion_integraiton}
\end{figure}

The virtual agent comprises of two key components: a speech LLM which handles spoken dialogue generation, and motion model which generates visual behavior. \Cref{fig:speech_lm_motion} shows a straightforward approach to the integration of motion model with speech model via the generated speech.
A more advanced approach is the integration via codebook as in \Cref{fig:speech_lm_code_motion}. Adapters are built on top of a speech LLM, taking its hidden states as input and predicting emotion and gesture codes. These codes serve as visual guidance for motion models besides the dyadic speech. The codebook integration is motivated by speech LLM's ability to grasp conversational context, enabling the extra controllability of generation of gestures and expressions that suit each moment of the conversation.

\begin{figure}[h]
    \centering
    \includegraphics[width=0.8\linewidth]{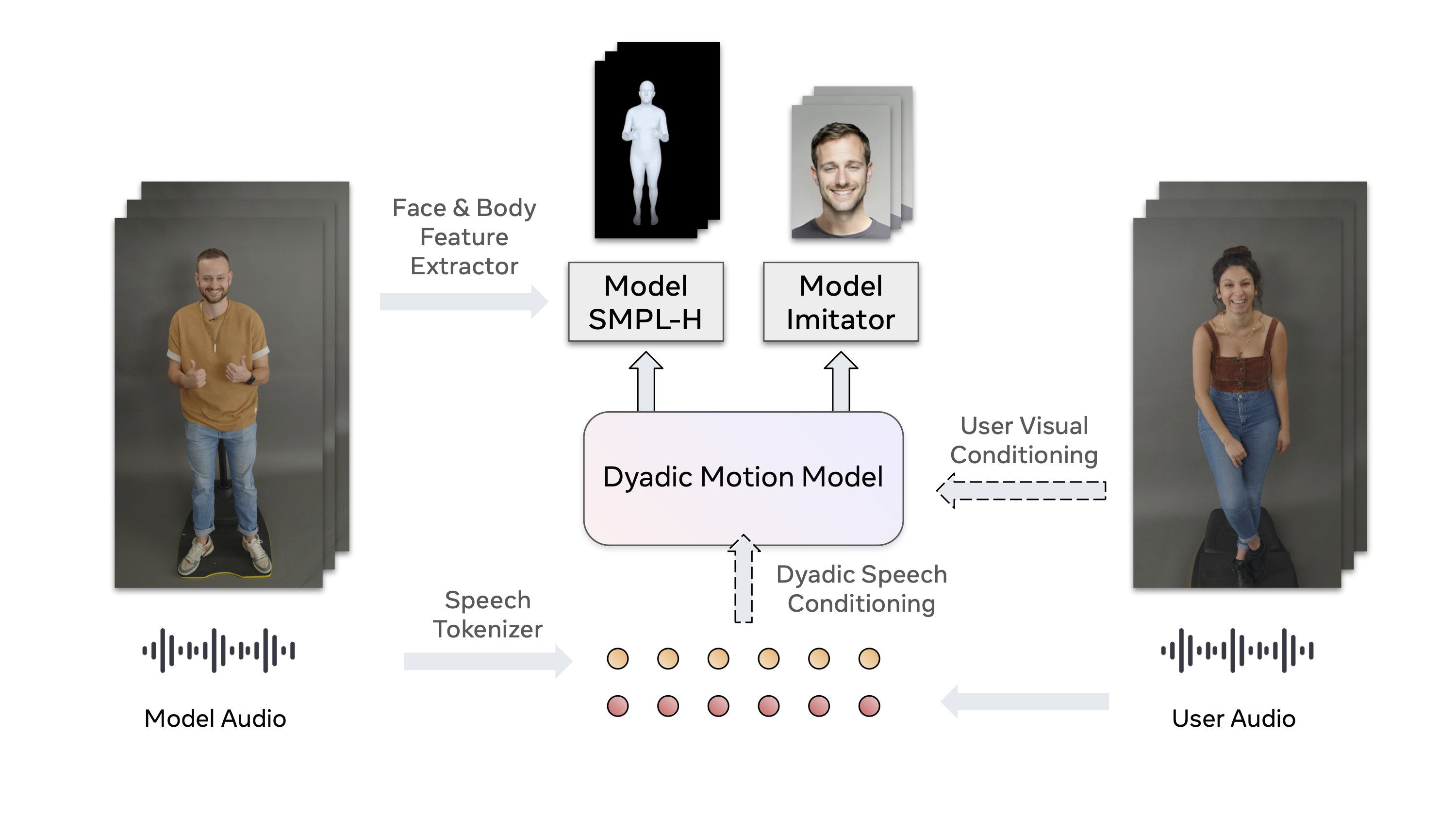}
    \caption{Dyadic Motion Model is conditioned on speech tokens obtained from dyadic audio and optionally user’s visual features, and generates face Imitator features and body SMPL-H features.}
    \label{fig:motion_model}
\end{figure}

We then share details of dyadic motion models as part of the virtual agent. Dyadic Motion Model in \Cref{fig:motion_model} is based on a Diffusion Transformer architecture trained with flow-matchin objective. The model is conditioned on dyadic audio input and optionally user’s visual features. 
There are controllable variants of motions models that can adapt emotion responses and expressivity levels, as well as generating more semantically-relevant gestures.
As for output, the motion model generates Imitator features as face representations, and SMPL-H  as body movement features.
These face and body features are compatible with both 2D and 3D renderings.

\subsection{Benchmarks}

The goal of virtual agents is to create realistic, engaging, and effective human-computer interactions. The key research topics are to design and evaluate virtual agents that can simulate human-like behavior, conveying emotions, intentions, and empathy through not only speech, but also facial expressions and body language.

% \Cref{fig:seamless_interaction_data} 
\paragraph{Dataset.} Seamless Interaction Dataset is vast and diverse collection of in-person, dyadic interactions designed to drive innovation in social AI research. With over 4,000 hours of interactions from more than 4,000 participants, this dataset provides an unparalleled wealth of data for researchers to explore. The dataset features nearly 1,300 conversational and activity-based prompts, covering a wide range of topics, interpersonal stances, and participant relationships. Anchored in contemporary psychological theory, the dataset includes both Naturalistic and Improvised content, offering a rich and nuanced understanding of human interaction. To further contextualize the interactions, the dataset is accompanied by detailed annotations and metadata, providing researchers with a comprehensive foundation for developing and evaluating social AI models. By leveraging this extensive dataset, researchers can advance our understanding of human behavior and develop more sophisticated and effective social AI systems.

\paragraph{Subjective metrics.} Evaluating behavior in embodied agents remains an unsolved problem, \cite{seamless_interaction} presents an approach to human subjective studies that incorporate face- and body-dyadic protocols. The face protocol focuses on facial expressions and head movements in photorealistic renderings. As for body protocol, it pays attention to visual behavior involving overall pose, hands, arms, shoulders, and head, but without facial rendering. The human studies use a pairwise approach where annotators view two dyads side-by-side, and participants provided a preference rating for each pair of visualizations. The protocols consist of $10$ core evaluative dimensions, which are grouped into three sections: overall preferences, listener-behavior preferences, and speaking-behavior preferences. The evaluation dimensions include key aspects such as lifelikeness, clarity of intent, turn-taking, listening, and speaking. By examining these critical dimensions, we can gain a deeper understanding of how well virtual agents simulate human-like behavior and interact with humans in a natural and effective way.

\paragraph{Objective metrics.} The research community have adopted automatic metrics to quantitatively evaluate the visual quality of face and body generations. In terms of face metrics, Lip-Sync scores including Sync-C and Sync-D evaluate the lip synchronization with speech, and Fr{\'e}chet Inception Distance (FID) to assess face image quality based on features encoded by the Inception network. For body generation, Fr{\'e}chet Gesture Distance (FGD) is widely used to quantify the discrepancy in distribution between generated outputs and real data. Furthermore, Diversity is also used to measure the range of variations present in the generated gestures.

\section{Type II: Wearable Agents }
\label{sec:type_2_wearable_agents}
Wearable devices represent a paradigm shift in human-computer interaction, standing distinctly apart from conventional smart devices due to their unique embodied nature. These devices are characterized by their integration of cameras, microphones, and an array of sophisticated sensors directly worn on the user's body, enabling them to capture an egocentric perspective of the physical environment surrounding the human user. This first-person perspective fundamentally transforms the relationship between artificial intelligence and human experience, creating what researchers term a ``shared perceptual field''.

The distinctive characteristic of wearable devices in the technological ecosystem stems from their ability to perceive the world from the user's vantage point. They can see what the user sees and hear what the user hears. Unlike stationary smart devices or even mobile phones that require explicit interaction, wearable AI systems can continuously monitor and interpret the user's environment through this egocentric lens. This continuous perception enables a form of ambient intelligence that remains contextually aware without constant user intervention.

Among the most significant implementations of wearable AI technology are Meta's AI Glasses \citep{waisberg2024meta}, which represent a convergence of advanced hardware capabilities and sophisticated AI systems. These glasses enable users to seamlessly access Meta AI through natural interaction methods, allowing them to request information, control smartphone applications, and engage in conversational interactions with the embedded AI assistant. The multimodal nature of the AI system empowers the glasses to perceive the user's visual field and auditory environment through specific voice commands or physical gestures, creating a contextually aware assistant.

\subsection{Capability Taxonomy}

A wearable agent, designed to assist and augment human capabilities, requires a range of advanced features to effectively predict and support user needs. Based on super sensing and perception, the agent should be able to anticipate which task the human user wants to perform next, using both explicit instructions and implicit contextual cues.
The agent's key capabilities include physical/mental dual world planning and execution, personalization, memory, social intelligence, digital tools and UI navigation, and physical tool/device/machine use guidance.
In terms of physical/mental dual world planning and execution, the agent should be able to model the physical environment and understand the user's interactions with it, as well as model the user's mental state, including their goals, preferences, and intentions. The agent should also be able to plan and execute tasks over extended periods, taking into account multiple factors and constraints, and adapt to changing circumstances by adjusting its plans accordingly.
The agent should be able to balance competing interests and optimize outcomes for all parties involved, reason about time and schedules, navigate the physical environment with ease, and plan and execute actions in the physical world using a range of tools and devices.
Personalization is another critical aspect of the agent's capabilities. It should be able to follow explicit instructions and high-level soft instructions from the user, adapt to changing user preferences and priorities, provide recommendations and suggestions based on the user's preferences and goals, seek out information in the physical surrounding, and follow timed and spaced instructions.
The agent should also have a persistent memory, allowing it to retain information over extended periods, recall specific events and experiences using episodic memory, and store and retrieve information over long periods using long-term memory.
Social intelligence is another essential capability of the agent. It should be able to pick up social cues from human conversations, understanding nuances and context, inform user action based on cultural and social cues, and improve conversation and engagement, reducing friction between humans.
In addition, the agent should be able to interact with Android agents and other digital tools, use Llama and other AI-powered tools to enhance its capabilities, provide guidance on the use of known tools and devices, adapt to new tools and environments using zero-shot learning, and access a large library of tools and devices to provide guidance on a wide range of tasks.
Overall, the wearable agent's capabilities should enable it to provide seamless and intuitive support to users, enhancing their productivity, efficiency, and overall experience.

\subsection{Architecture and Models}
\label{sec:wearable_architecture_models}

\begin{figure}
    \centering
    \includegraphics[width=0.85\linewidth]{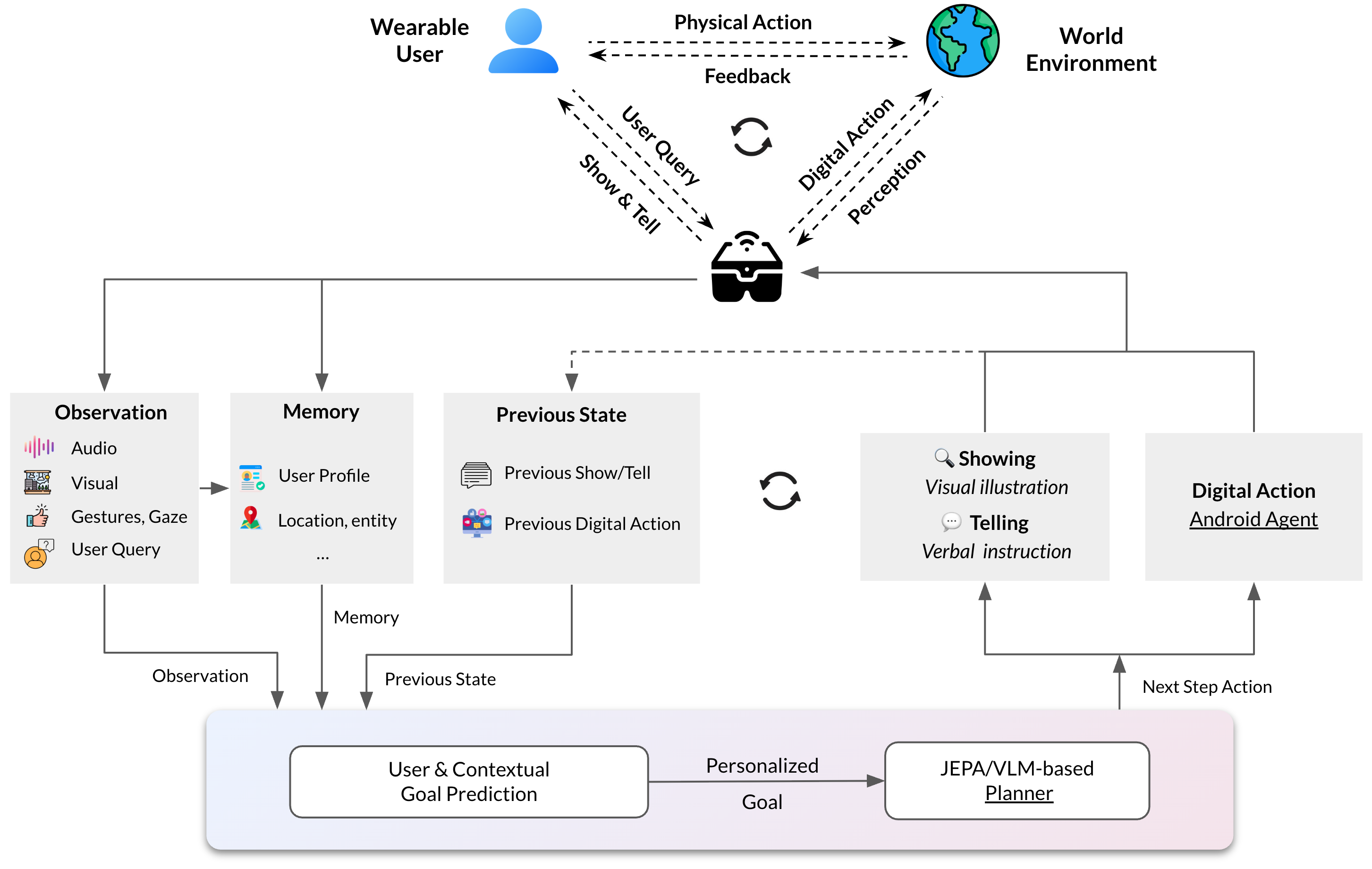}
    \caption{Wearable agent architecture to propose the next action step using a contextual goal prediction module and a planner.}
    \label{fig:enter-label}
\end{figure}

High-level world modeling depends critically on designing predictive architectures that reason over semantically meaningful transitions. These models must not only forecast future states but do so in a way that supports compositionality, causality, and goal-directedness. While generative vision models can simulate pixel-level future frames, they are ill-suited for abstract reasoning due to their computational cost and inability to represent high-level semantic changes compactly. Instead, predictive models that operate in latent or linguistic abstract spaces provide a more scalable and generalizable alternative. These models can forecast outcomes of actions without reconstructing unnecessary sensory detail, making them ideal for high-level planning tasks.

This motivates a turn toward self-supervised learning from large-scale unlabelled video corpora, particularly from egocentric and instructional recordings found in kitchens, workshops, and clinical settings. Such videos naturally contain rich sequences of human activity, including implicit goals, actions, and state transitions. A high-level model trained on such data should be able to segment and identify meaningful steps in the procedure, infer underlying intentions and causal relationships between different steps, predict subsequent actions and possible states under diverse assumptions, and represent the task structure in an interpretable and easily conveyable to human users.

Vision-Language World Model (VLWM) is a predictive model trained on unlabelled video data to generate interleaved natural language sequences that describe actions and resulting world states. These textual representations serve as both output and planning substrate—interpretable, compositional, and accessible to large language models. VLWM is trained to condition on visual context and generate plausible procedural futures in language, without requiring annotated supervision.
This allows VLWM to simulate future trajectories, support candidate plan evaluation, and reason over causal dependencies—all in a symbolic space that aligns with human task understanding. Empirical evaluations using Visual Planning for Assistance (VPA) benchmarks show that VLWM significantly outperforms prompting-based baselines, achieving relative improvements of +20\% in success rate (SR), +10\% in mean action accuracy (mAcc), and +4\% in mean intersection-over-union (mIoU). Human evaluation via PlannerArena further confirms that plans generated by VLWM’s system-2 reasoning mode are consistently preferred over alternatives.

Another architecture which would be meaningful to explore are JEPA-based planners. These models would also be conditioned on high-level prompts (e.g., textual task descriptions) but forecast outcomes directly in latent video representation space rather than language. While less semantically transparent than VLWM, JEPA-based planners offer two key advantages: trajectory optimization via differentiable planning objectives, allowing for gradient-based refinement, and test-time efficiency, especially when integrated with model predictive control (MPC) for fast online decision-making. This approach trades off interpretability for performance, offering a complementary pathway for high-level reasoning in agents that must operate under resource constraints or require rapid adaptation. By systematically contrasting VLWM and JEPA-based planners, we can clarify the trade-offs between linguistic expressivity and optimization-driven control, guiding future designs of embodied planning systems.

\subsection{Benchmarks}

\subsubsection{Goal Inference Benchmark}

\begin{figure}
    \centering
    \includegraphics[width=1\linewidth]{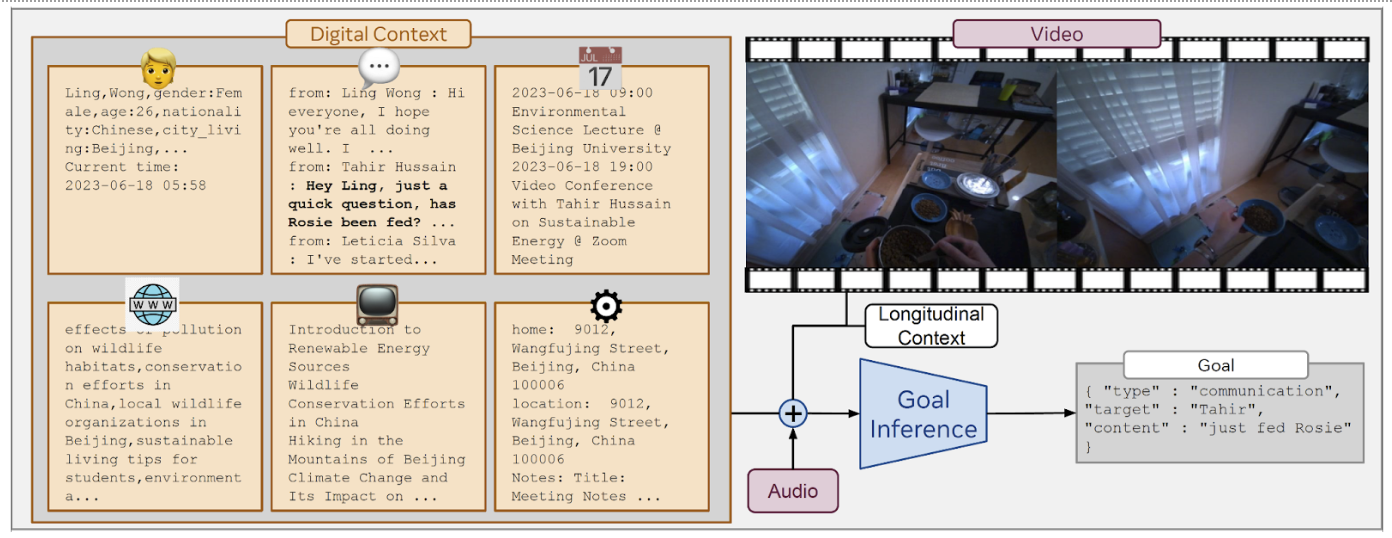}
    \caption{A multi-modal sample from the goal inference benchmark. Here the video and digital contexts are relevant to the prediction problem, and audio/longitudinal are noise.}
    \label{fig:enter-label}
\end{figure}
Agentic systems have the potential to be very powerful tools, but if each query to such agents require exhaustively specific requests the queries become as onerous to generate as performing the task by oneself. To this end, the Egocentric Multi-modal Goal Inference Benchmark looks at the problem of inferring one's goal from multi-modal contextual observations. Solving this `goal inference' problem holds the promise of eliminating the effort needed to interact with such an agent. This work focuses on creating a strong benchmark to measure progress in solving this problem using vision-language models (VLMs). Given the limited prior work in this area, the team collected a novel dataset consisting of multimodal data from 348 participants across 3,477 recordings, featuring ground-truth goals. Figure below shows the diversity of modalities present in this benchmark, which for the first time brings together visual, audio, digital, and longitudinal contextual observations required for inferring the wearer’s goal.

Human performance exceeds model performance, achieving 93\% multiple-choice accuracy compared with 84\% for the best-performing VLM. Benchmark results that evaluate several families of modern vision-language models show that larger models perform significantly better on the task, yet remain far from practical usefulness, as they produce relevant goals only 55\% of the time in the generative setting. Through a modality ablation, it is shown that models benefit from extra information in relevant modalities with minimal performance degradation from irrelevant modalities.

\subsubsection{WorldPrediction Benchmark}

\begin{figure}[h!]
    \centering
    \includegraphics[width=0.7\linewidth, trim=0 80 0 80, clip]{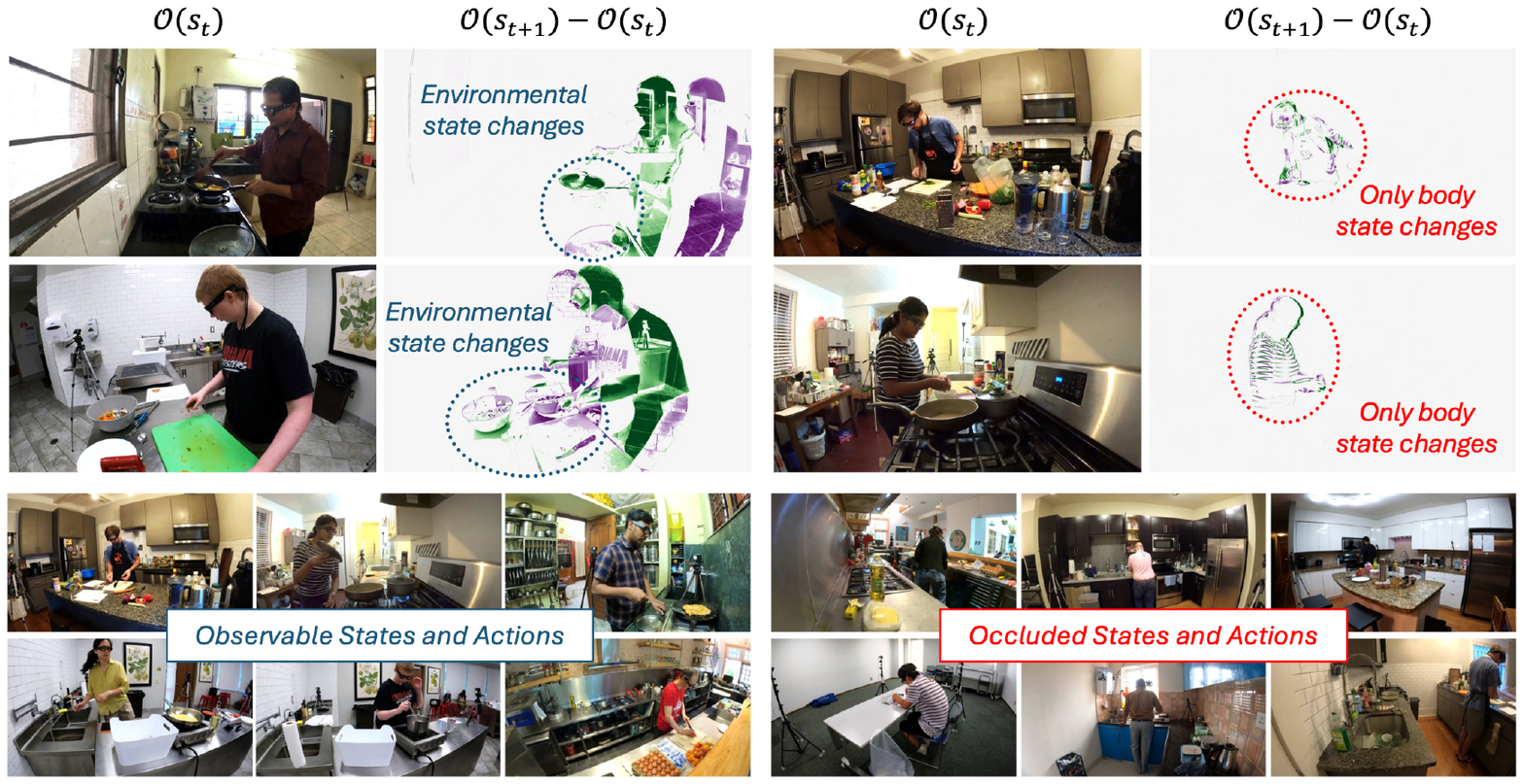}
    \caption{Observable vs. Occluded high-level states and actions in WorldPrediction. We show the environmental and body state changes between initial and final states taken from EgoExo4D \citep{grauman2024ego}.}
    \label{fig:world_prediction_actions}
\end{figure}

We refer to the underlying details of WorldPrediction in section \ref{sec:world_prediction}. This benchmark also serves as a practical touchstone for the specific challenges faced by wearable agents. Egocentric devices must transform a rapidly shifting first-person video stream into predictive representations that guide on-device assistance (e.g., suggesting the next tool while cooking or flagging a safety hazard before it occurs). Because its two tasks require choosing high-level actions and full action sequences under partial observability, WorldPrediction mirrors the temporal abstraction and uncertainty inherent to on-body perception. Moreover, its “action-equivalent” design forces models to rely on inferred causal structure rather than pixel-level continuit, exactly the type of abstraction a resource-constrained wearable must compute to remain responsive. WorldPrediction also includes egocentric action segment with exocentric state observations from EgoExo4D as shown in figure \ref{fig:world_prediction_actions}. Thus, evaluating an agent’s policy generator on WorldPrediction yields a direct measure of whether it can anticipate user intent several steps ahead, a prerequisite for proactive, context-aware assistance in open-world settings.

\section{Type III: Robotic Agents }
\label{sec:type_3_embodied_agents}
Robots are the quintessential embodied AI agent. They are typically equipped with sensors (e.g., RGB cameras, proprioception, touch) to perceive and actuators to interact with the environment. While the term robot encompasses systems commonly used as advanced tools (e.g., industrial or medical surgery robots), we focus here on robots designed to operate in everyday settings (e.g., homes or offices) for general purpose tasks with some degree of autonomy. The promise of such systems is to help people with everyday tasks. Recently, improvements in robot intelligence have led to systems that can be trained with human demonstrations (i.e., recordings of people teleoperating the robot) to accomplish tasks such as folding laundry or picking-up objects to clean a room \citep{intelligence2025pi_,bjorck2025gr00t,team2025gemini}. However, these capabilities are often limited to settings similar to those collected in the training data. Thus, we are yet to unlock the full potential of these systems to provide an extra set of hands to take care of any task that a user requires.

\subsection{Capability Taxonomy}

A robot's capabilities can roughly be divided into two abstract categories the ``inherent'' \emph{physical capabilities} of the robot - which is advanced by progress in hardware design, sensing, learning of motor-control for manipulation and locomotion, utilizing multi-fingered hands, dexterity, etc - and the \emph{brain} of the robot - which is advanced by progress in reasoning, planning, semantic understanding, memory, generalization, learning from humans, interacting with humans and lifelong learning. And while we list capabilities in either category separately, we do not mean to imply that they should be developed completely ``independently''.

\subsubsection{Physical Capabilities}
The physical capabilities of generalist robots are often categorized into three bins: locomotion, navigation, (dexterous) manipulation. 

\paragraph{Locomotion and navigation:} Both, locomotion and navigation, focuses on the movement of the robot from point a to point b; ideally across any terrain. Locomotion via legged robots over uneven surfaces and avoid small obstacles along their path is a particularly challenging problem that has seen tremendous progress in recent years. By contrast, navigation generally involves (implicitly or explicitly) planning a series of movements to reach a goal specified in either explicit 3d coordinates or through visual or textual goal descriptions (e.g., ``find and go to the kitchen from the living room''). 

\paragraph{(Dexterous) manipulation} Manipulation involves any purposeful interaction with objects such as generalized grasping and placing of objects, as well as advanced and meaningful tool use. While basic capabilities of grasping (and to some extend) placing have made significant progress, we're yet to see an agent that can perform truly general pick\&place. An important sub-category of manipulation is dexterous manipulation, which involves finegrained control of objects (e.g., ``put the key in the lock'', ``re-orient object in hand''), often requiring or benefiting from using multi-fingered hands. This area is even more challenging than the more coarse picking and placing manipulation skills, and is still an open research problem. 

While most research focuses on one of the physical capabilities, there is an increased focus on learning policies for ``full body control'' which attempts to learn policies that can control both the location of the robot in it's environment as well as the manipulation of the environment.

\subsubsection{``Brain'' Capabilities}
In addition to the motor control capabilities, general purpose robotics also requires more higher-level intelligence and reasoning capabilities as well. 

\paragraph{Generalization.} A robotic agent needs to be able to generalize its knowledge and skills to new scenarios. For instance, just like humans, robotic agents should be able to learn how to make a certain dish in one kitchen, and then be able to repeat this task in new kitchen (perhaps because the robot and human had to move into a new house). Generally speaking, there are multiple axes of generalization that we hope a robotic agent can master: Generalization to new tasks and skills (action generalization), to new embodiments and to new environments and objects (visual/semantic generalization).
    
\paragraph{Efficient and lifelong adaptation.} Even with the best generalization capabilities, its infeasible to prepare a robot for everything that might be needed at deployment time. A robot must be able to efficiently adapt when it encounters unfamiliar scenarios without forgetting previously learned skills and capabilities. For instance, while a robot might have already learned how to fold clothes in some way, a user may want their clothes folded in a specific other way. This type of personalization should not require many data samples, should not lead to forgetting of other related skills (folding towels) and ideally can be done by a user simply (visually) demonstrating their desired clothes folding procedure.

\paragraph{Spatial and temporal memory.} 
A robotic agent ultimately has to operate over long time horizons (perhaps executing on one very long task, or simply because of executing many smaller tasks). Because of this the agent needs to build up a spatial memory (e.g. through semantic mapping) of where important parts of the living space are (bed room, ...) as well as where objects tend to be located. It also has to be able to keep track of what sub-tasks/actions it has already executed when performing a task such as cooking which involves many steps. Finally, a robot should be able to retain important previous user interactions (e.g., personalization).

\paragraph{Language instructions and planning.}
A robotic agent must be capable of following language instructions at multiple levels of abstractions and must be able to break down such task requests into smaller units of activities. For instance, when asked to perform the task of ``clean up the kitchen'', the agent needs to semantically understand the request, figure out what sub tasks to perform (``find dirty dishes'',  ``bring dirty dishes to dishwasher'', ``empty dishwasher first if needed'', etc.) and then plan both at that abstract task level as well as at the more concrete sub-task level in dynamic environments. 

\paragraph{Interaction with humans and other agents.} Last, but not least, a robot will be interacting with humans or other robots. Interaction with users involves receiving instructions, asking clarifying questions, and receiving corrections. Another desired interaction would be for a robot to physically help a human to perform a task (such as needed in elderly care). Finally, it can also mean collaboration with a human (or other robot) on a task and having to divy up responsibilities for a task. In some of these interactions, it's important for the robotic agent to be able to infer the intent of other agents in the scene (e.g. through building a mental world model).

\subsection{Architecture and Models}

Skill and task learning in robotics remains a very active and challenging research domains. Many different approaches exist. Here we primarily pick two very contrasting approaches to help motivate the role world models can play.

Recently, Vision-Language Action Models (VLAs)\citep{intelligence2025pi_,bjorck2025gr00t,team2025gemini,shukor2025smolvla} trained from teleoperation data have emerged as a promising route towards general purpose robotic foundation models. These models leverage VLMs to infuse policies with general world knowledge with language-conditioning which (in theory) allows for open-set task specification. Thus far these types of approaches have been focused on generalize pick and place behaviors with the goal of generalizing to new environments and objects. High-quality data generation (clean teleoperation data) is essential to this line of work and remains one of the main bottlenecks.

Alternatively, policies trained via reinforcement learning (RL) in simulation with a focus on sim2real transfer, continue to be a promising route. In particular, deep reinforcement learning (DRL) has proven successful for learning full-body controllers of humanoids for locomotion and navigation tasks, and some dexterous manipulation tasks. Thus far progress has been limited to tasks that we can easily specify reward functions for. The main bottlenecks for this line of work remains reward specification, sim2real transfer (due to limited scene variability and imprecise object interaction physics of the simulators) and simply scaling RL to multi-task policies with language-conditioned task specification.

\begin{figure}
    \centering
    \includegraphics[width=0.49\linewidth]{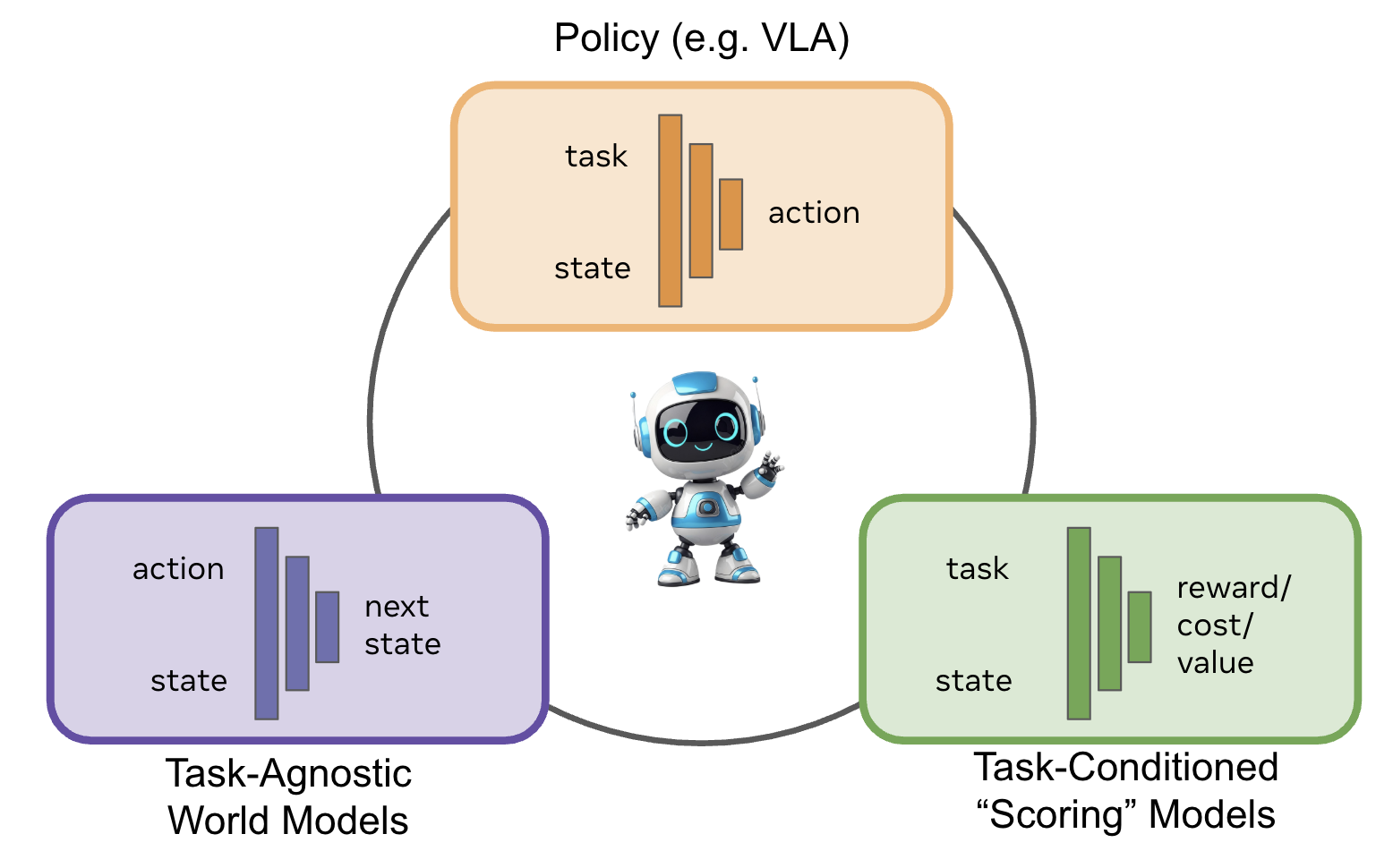}
    \includegraphics[width=0.45\linewidth]{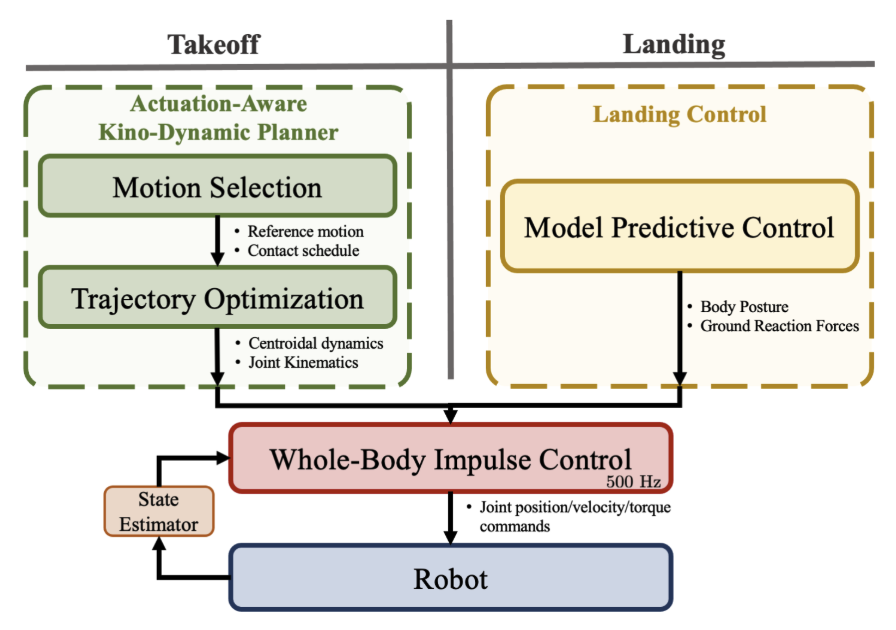}
    \caption{({\bf left}) We envision a framework that combines policies, task-agnostic world models and task-conditioned cost/scoring models. For instance, the world model + scoring model can detect if the VLA would fail with its predicted actions and re-plan based on the agents world model  ({\bf right}) Example of nested control loops utilizing analytical models at various frequencies for a complex tasks such as the MIT Humanoid Robot performs a backflip jump (Taken from \cite{MITHumanoidRobot2021}).}
    \label{fig:embodied-system}
\end{figure}

%~\cite{DeepRLSurvey}
Both of these approaches inherently assume that generalization emerges as a function of data scale (whether teleoperated or through interaction in the simulator) \emph{and} that a robot agent can ``learn once everything the robot needs to know''. However it is entirely unclear whether it is feasible that we can prepare the robot for everything it might encounter in the real world, or whether we can even learn a policy that can represent all possible skills in every possible scenario. Instead of trying to do so, we should also enable the robot to understand how its actions affect the world (through learning a world model), such that it can find new solutions at test (inference time). World models are a promising route to achieving general robot behaviors with limited in-domain data, and naturally enable training from various (heterogeenous) data sources (successful and unsuccessful task executions, video data, play data). Ultimately we believe that a system combining policies, world models and ``reward'' models will be our best bet to general purpose robotic agents (see Figure~\ref{fig:embodied-system} (left)).

\subsubsection{A Note on Classical Robotics utilizing Analytical Models}
\label{subsubsec:classical-robotics}
To enable extract extractions from a world model we need to perform reasoning/planning with a world model, and we imagine to do this similar to how model predictive control is performed. Model predictive control with (analytical) world models has a long and rich history in robotics, especially for ``lower'' level control of robot motions (see Figure~\ref{fig:embodied-system} for an example). Although here we discuss learned (action-conditioned) world models (and discuss world models at different levels of action abstractions), it is not necessary to reinvent the wheel on the axis of planning with world models. Instead we can leverage and/or build on advances in robotic control with analytical models.

To perform low-level control tasks, such as locomotion on uneven terrain~\cite{Nguyen-dynamic-walk}, climbing obstacles~\cite{Anymal}, jumping~\cite{Jumpinglegged2022} or even performing a backflip with a humanoid robot~\cite{MITHumanoidRobot2021}, the field of robotic control has classically relied on kyno-dynamic planning~\cite{Planning_Algorithms_book}, Model Predictive Control (MPC)~\citep{borrelliPredictiveControl} and an analytical physical model of the robot. Most of these models do not integrate any deep learning component, as they leverage hardcoded knowledge about the body’s and the environment’s dynamics. They rather rely on control theory and the physical equations of motion at various levels. The example of the MIT Humanoid Robot~\cite{MITHumanoidRobot2021} show-cases these different analytical control approaches within a hierarchical framework to allow the humanoid robot to perform a backflip jump, as illustrated in \cref{fig:embodied-system}. 

\subsection{Benchmarks}

The ultimate goal of robotics is to deploy systems into the real world. Thus, real-world evaluations are the “gold standard.” However, evaluating robots in the real world is both time consuming and difficult to do in a reproducible way. Thus, three types of benchmarks are used in robotics, trading-off realism, reproducibility, and scale:

\emph{Offline benchmarks} leverage real-world data to evaluate specific capabilities of robots such as the ability to identify an object in a 3D scene. While such benchmarks are reproducible and scalable, they do not evaluate the end-to-end performance of a model, system  or agent on a robot. \emph{Simulation benchmarks} leverage physics engines (such as MuJoCo or pyBullet), and are fully reproducible. However, these simulations are often not photorealistic and struggle to accurately model finegrained interactions between robots and objects. \emph{Real-world hardware evaluations} are often conducted at small scales within lab environments and are generally not reproducible. Furthermore, very few public hardware evaluations exist. In the following we focus on end-to-end evaluation of learning based approaches.

\subsubsection{Simulation Benchmarks}

Simulation benchmarks~\citep{ManiSkill2, Metaworld, DMControl, james2019rlbench, LIBERO, robocasa2024, Habitat3.0, Habitat2, xiazamirhe2018gibsonenv,ai2thor,simpler, RoboHive}, provide a fully reproducible setup to efficiently evaluate robotic agents. However, simulation benchmarks often either lack photorealism and/or may not properly model the physics of interaction with objects or the movement of deformable objects (e.g., clothing). This can make it hard to either transfer models directly to robot, or to do model selection in simulation as it is unclear whether simulation performance is predictive of hardware performance. For some capabilities, such as visual navigation, simulation benchmark performance has positively correlated with real-world metrics ~\citep{kadian2020sim2real, Silwal2024sim2real} especially when utilizing encoders pretrained on real data - such as VC-1~\citep{majumdar2023we} - in both sim and real experiments. However, for manipulation tasks thus far mixed results have been found when studying correlation between simulation and hardware performance~\citep{Silwal2024sim2real}. This is presumably the case because for manipulation the interaction with the environment matters more (as compared to navigation or locomotion tasks).

\subsubsection{Hardware Benchmarks}

Due to the challenges of setting up and maintaining a robot evaluation environment, large-scale public robot benchmarks that evaluate generalist robots across diverse environments do not exist. However some explorations in this direction exist. Challenges like the HomeRobot Challenge~\citep{yenamandra2023homerobot} provide a simulation platform for researchers to prototype and test their algorithms, and then invite best performers to also test their algorithm on hardware. There are also first attempts at providing an ``evaluation on hardware service'' that allows researchers to submit policies that then get evaluated either fully automatically through learned success classifiers~\citep{zhou2025autoeval}, or through deploying human evaluators that compare and rank policies~\citep{atreya2025roboarena}. These explorations are in its infancy, and many questions remain unanswered, such as how to robustly extract metrics in the non-stationary environment (robot and environments change over time) these evaluations are performed on.

\section{Future Direction: Embodied AI Learning }
\label{sec:future_direction_embodied_ai_learning}

\begin{figure}
    \centering
    \includegraphics[width=0.75\linewidth]{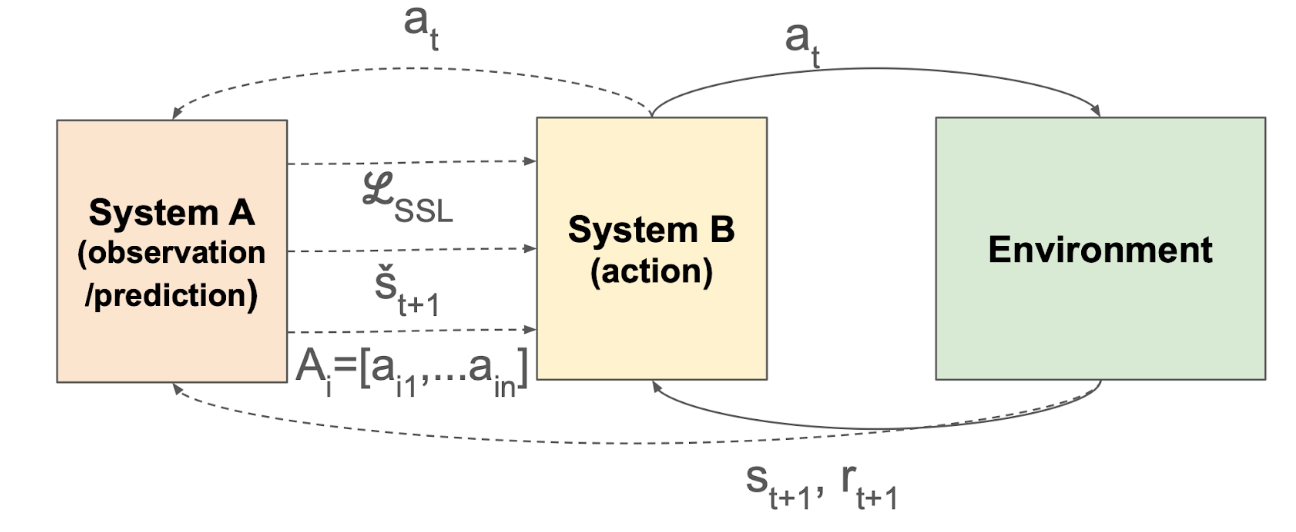}
    \caption{\textbf{Summary of modes of interactions between Systems A and B}. System A provides System B with predictions of future states (st-1) conditioned on past states and actions, with hierarchical abstractions over possible actions (Ai), and a SSL loss that can be used for curiosity/exploration. System B through its action provides rich and task relevant input for System A to learn from.  }
    \label{fig:7-system-a-b}
\end{figure}

To achieve autonomous learning, AI systems must integrate passive perception and active behavior while shifting from datapoint-level to task-level generalization, enabling fast adaptation through self-supervised task discovery and interaction-driven learning.

In contrast to artificial systems, animals learn and act from birth \citep{smith2005development}. Learning is continuous, interactive, and goal-directed, without being siloed into distinct training phases or frozen once completed. Current AI systems, by comparison, separate learning and action into distinct paradigms (e.g., self-supervised learning, reinforcement learning), each requiring heavily engineered data pipelines and sequential recipes (e.g., pretraining followed by fine-tuning). Learning often ceases altogether once models are deployed. In this paper, we argue that achieving autonomous learning will require addressing the architectural integration of the different paradigms into a system that can both learn and act.

One can sort out the different learning algorithms into two broad categories or paradigms: observation-based learning and action-based learning, which we will call System A and System B, respectively. System A extracts structure and patterns from passive sensory data. System B interacts with the environment to drive learning through goal-directed behavior. While both paradigms have shown impressive progress independently, they each have fundamental limitations when used in isolation. 

\paragraph{System A: Learning by Observation. } System A comprises learning mechanisms that extract abstract representations from raw sensory inputs. This includes self-supervised and unsupervised learning models applied to static datasets or passively collected sensory streams \citep{oord2018representation,nguyen2020deep,devlin2019bert}. These paradigms can be classified based on their modality, data type, and structure. Some systems operate on a single modality such as text, images, or audio, while others combine modalities, for instance vision and language \citep{radford2021learning}. Some work on symbolic, discrete data like tokens, while others learn directly from continuous sensory input. Finally, some models explicitly exploit the spatial or sequential structure of their inputs, such as grids or time series.
System A has several strengths. It scales well with large datasets and is capable of discovering abstract latent representations that can be organized hierarchically across different levels of abstraction, from low-level sensory features to high-level conceptual categories \citep{bengio2013representation}, and that can support transfer to downstream tasks \citep{devlin2019bert,caron2021emerging}.
However, it also faces limitations. These systems need access to large and well-curated datasets, or long periods of passive data collection. They lack any built-in mechanism to decide what data might be useful or should be acquired next. Furthermore, their representations are disconnected from the agent's ability to act, making it hard to ground what they learn in real-world behavior. Because they are based purely on observation, they find it difficult to distinguish between correlation and causation.

\paragraph{System B: Learning by Action. } System B comprises learning mechanisms that operate through interaction. The most common paradigm is reinforcement learning, where an agent explores an environment and learns to select actions that maximize a reward \citep{Sutton1998}. These systems vary in their reward sources, which may be provided externally through task-specific signals or generated internally through curiosity, novelty, or empowerment \citep{schmidhuber1991learning, pathak2017curiosity,rezende2015variational}. The size and structure of the action space also varies widely. Simple environments use a small set of discrete actions, but real-world tasks often require complex, continuous, and high-dimensional action spaces \citep{lillicrap2015continuous}. Exploration strategies can be random, curiosity-driven, or guided by a goal or policy \citep{ecoffet2021first}.
System B has important strengths. It is grounded in control and interaction, enabling it to learn directly from sparse or delayed outcomes. It is naturally suited for real-time and adaptive behavior. It can also discover truly novel solutions by search. But it also faces major limitations. It is sample inefficient, often requiring large numbers of interactions to learn even simple tasks. It struggles in high-dimensional or open-ended action spaces. And it depends on having well-specified reward functions and interpretable actions, which are often unavailable in naturalistic settings.

\paragraph{System A Helping System B. } The integration of System A and System B could address several limitations of using System B alone. Because System B suffers from sample inefficiency and intractability in high-dimensional spaces, System A can support it by providing structure, priors, and compressed representations that make learning and planning more tractable.

Each of these mechanisms reduces the burden on System B by filtering, structuring, or simulating the environment in abstract spaces learned by System A. What remains to be done is to propose an integrated architecture that can support these different modes of learning within a single system applied for instance to large amounts of video data and small amounts of aligned video and action. A possible North Star for this would be a robot that would learn, like infants do, useful action / vision world models from observation of its video stream plus motor babbling \citep{dwibedi2018learning}

\paragraph{System B Helping System A. }
System A's primary limitation is its reliance on passive data. Without guidance, it may learn from uninformative or irrelevant data streams. System B, through active behavior, can help solve this problem by collecting better data and providing grounding for learned representations. Gibson’s \citep{gibson1966problem} notion of active perception: “We see in order to move and we move in order to see” is a classic statement of the active gathering of information driven by a goal.

System B can support System A in two fundamental ways: directly, by optimizing System A’s predictive objectives, or indirectly, by exploring the environment in ways that yield task-relevant or informative trajectories. In both cases, the help provided to System A can take two forms: the generation or selection of useful data, and the creation of enriched data in the form of parallel corpora consisting of paired actions and sensory consequences \citep{hill2020human,lynch2020learning} 

In both cases, System B enriches the data available to System A by generating parallel action/perception datasets that support cross-modal learning or even a form of supervised learning, to the extent that actions are lower-variance and more reliable than noisy, ambiguous sensory inputs.

In brief, System B has the potential to remove the data preparation and filtering step that is currently extremely resource intensive, by directly extracting the data from the outside environment. Of course, this assumes that it is possible to interact with the environment at scale, either through embodied agents or large-scale simulations.

Together, these bidirectional pathways lay the foundation for a learning system that can act and learn in tandem, where perception informs action and action fuels perception, see Figure~\ref{fig:7-system-a-b}.

\begin{table}
  \centering
  \caption{Examples of System A + B Integration in Machine Learning.}
  \label{tab:system-ab}
  \renewcommand{\arraystretch}{1.25}
  \setlength{\tabcolsep}{6pt}

  \resizebox{\textwidth}{!}{%
  \begin{tabular}{|l|p{3.5cm}|p{3.8cm}|p{4.2cm}|p{3.2cm}|}
    \hline
    \textbf{Domain} &
    \textbf{System A (Observation)} &
    \textbf{System B (Interaction)} &
    \textbf{Resulting Learning Loop} &
    \textbf{References} \\ \hline\hline

    % ------- rows -------
    Motor Control &
    Predictive modeling from video (e.g., \textit{Dreamer}, \textit{Video-JEPA}) &
    Policy learning in simulation (e.g., RL with imagination) &
    World models enable sample-efficient control via latent imagination &
    Hafner \textit{et al.}, 2019, 2020; Bardes \textit{et al.}, 2024 \\ \hline

    Language Learning &
    Pretraining on large-scale text (e.g., next-token prediction) &
    Reward shaping via human feedback (e.g., RLHF) &
    System A learns linguistic representations; System B aligns them with human preferences through interaction &
    Ouyang \textit{et al.}, 2022; Bai \textit{et al.}, 2022 \\ \hline

    Complex Skills &
    Representation learning from unlabeled play data &
    Goal-conditioned behavior learned from demonstrations &
    Self-supervised embedding of latent plans and affordances from behavior &
    Lynch \textit{et al.}, 2020 \\ \hline

    Social Learning &
    Learning from observing human interactions in multi-agent settings &
    Interactive probing to infer intent or beliefs &
    Not done to our knowledge & -- \\ \hline
  \end{tabular}
  }
\end{table}

Implementing this framework in practice requires both architectural and training innovations. Pretraining should be evaluated not just in terms of performance on fixed benchmarks, but in terms of how quickly and effectively the system can learn new, unseen tasks. This calls for new benchmarks and training paradigms centered around task distributions, curriculum learning, and the automatic discovery of goals or subtasks. The integration of SSL and RL within the pretraining loop also calls for massive investment in simulators that can deploy a vast array of procedurally generated tasks and environments.

\section{Future Direction: Multiagent Interactions}
\label{sec:future_direction_multiagent}
%\paragraph{Multi-agent interactions for embodied AI. }
% 
When multiple embodied AI agents work together, they can achieve complex tasks that would be difficult or impossible for a single agent to accomplish. This synergy occurs in multiple domains. One example is multi-robot systems, where a team of robots collaborate in a disaster relief scenario; one robot could assess structural damage, another can search for survivors, and a third robot could deliver medical supplies. By working together, each robot with its unique capabilities contributes to the overall mission, exceeding what a single machine could do. 
Similarly, a fleet of autonomous vehicles operating in a city requires sophisticated multi-agent collaboration, with vehicles communicating their paths, negotiating intersections, and adapting to traffic changes to ensure safe and efficient transit. 

In the realm of wearable devices, a user might wear smart glasses, a smartwatch, and a haptic feedback vest. These agents, embodied within the user, can collaborate to provide a richer, more integrated experience. For example, the smart glasses may identify an object, the smartwatch could pull up relevant information, and the haptic vest could guide the user's attention. 

In addition, in immersive virtual and augmented reality environments, various virtual embodied agents, such as personal assistants or non-player characters and user-controlled avatars, can engage in complex collaborative activities such as building structures or participating in shared quests, with research exploring their integration in augmented reality settings \citep{wang2019exploring}. 

Finally, a family of embodied agents might need to work together to help humans. For example, one can plan a dinner with friends by asking an agent to recommend menus,  go grocery shopping and cook the meal while being guided by a wearable agent. A robotic agent can assist in the kitchen and clean up after the dinner. Effective collaboration among diverse embodied AI agents requires careful consideration of fundamental factors, including communication, coordination, and conflict resolution.

\paragraph{Key Challenges. }
Enabling seamless multi-agent collaboration in embodied AI presents significant hurdles. Communication is crucial; agents must be able to share information and coordinate their actions. This extends beyond mere data transmission to establishing robust and meaningful communication protocols in noisy, stochastic or unpredictable environments. One challenge lies in how agents may convey intentions, share perceptions, and understand each other's messages, especially as they may possess differing sensory capabilities or even need to develop their own communication signals \citep{foerster2016learning}. 

Coordination is also vital, as agents need to synchronize their actions and overcome conflicts so as to achieve shared goals. This includes decentralized control, ensuring actions are executed at the right time, allocating bounded resources, and coordinating when agents have only a partial view of the overall environment \citep{weiss1999multiagent}. 

Finally, conflict resolution is an unavoidable necessity. Situations will inevitably arise where agents' goals or actions become incompatible. The ability to resolve these conflicts and negotiate is essential for sustained collaboration \citep{cao2018emergent}, encompassing instances where individual sub-goals clash with the collective objective, or when agents physically interfere with one another. Developing mechanisms for agents to negotiate and compromise without explicit human intervention remains a complex problem, often explored through frameworks designed for mixed cooperative-competitive environments \citep{lowe2017multi}. A fundamental requirement for effective cooperation is that machines must learn to find common ground in their understanding and actions \citep{dafoe2021cooperative}.

Existing foundations that could be leveraged: the field of multi-agent systems offers valuable insights and methodologies for addressing these challenges. We can leverage foundational understanding of how distributed AI agents communicate, coordinate, and resolve disputes \citep{weiss1999multiagent}. Research into emergent communication \citep{foerster2016learning} provides methods that enable agents to learn to develop their own communication protocols, to foster more autonomous and flexible embodied systems. 

Further, negotiation-based approaches (Meijer \& Tuyls, 2019) provide valuable frameworks for how agents can learn to negotiate and reach agreements, moving beyond simplistic rules for conflict resolution. 

Advancements in multi-agent reinforcement learning \citep{nguyen2020deep} offer powerful tools for applying deep reinforcement learning to complex multi-agent scenarios, especially involving both cooperative and competitive elements, enabling agents to learn optimal policies through dynamic interaction. The necessity for cooperative AI machines to find common ground in their interactions and goals is also a crucial area of study \citep{dafoe2021cooperative}. 

Surveys on cooperative heterogeneous multi-robot systems also offer valuable insights into effective collaboration strategies \citep{rizk2019cooperative}, while work on virtual agents provides context for their use in augmented reality and metaverse environments \citep{wang2019exploring}.

Best Practices: using these insights we can offer several best practices that can guide the design of effective multi-agent embodied AI systems. A shared goal or reward function enables us to align agents' objectives, incentivizing cooperative behavior, and encouraging them to work towards a collective outcome. Implementing a clear and efficient communication protocol is equally essential; this could involve predefined messages or more advanced, learned communication schemes tailored to the agents' capabilities and environmental context. 

Employing an appropriate coordination mechanism is crucial for synchronizing agents' actions and preventing interference, ranging from centralized controllers for smaller systems to distributed algorithms for larger, more flexible deployments, as often explored in surveys of heterogeneous multi-robot systems \citep{rizk2019cooperative}. 

Finally, integrating robust conflict resolution mechanisms is vital for maintaining collaboration in the face of inevitable disagreements. These mechanisms might include negotiation protocols, arbitration systems, or learning-based approaches where agents learn to avoid or mitigate conflicts through experience. By thoughtfully applying these best practices and leveraging the extensive research in multi-agent systems, we can design and implement increasingly sophisticated systems that enable embodied AI agents to work together seamlessly, unlocking new frontiers in AI capabilities and real-world applications.
\begin{figure}
    \centering
    \includegraphics[width=0.5\textwidth,height=0.15\textheight]{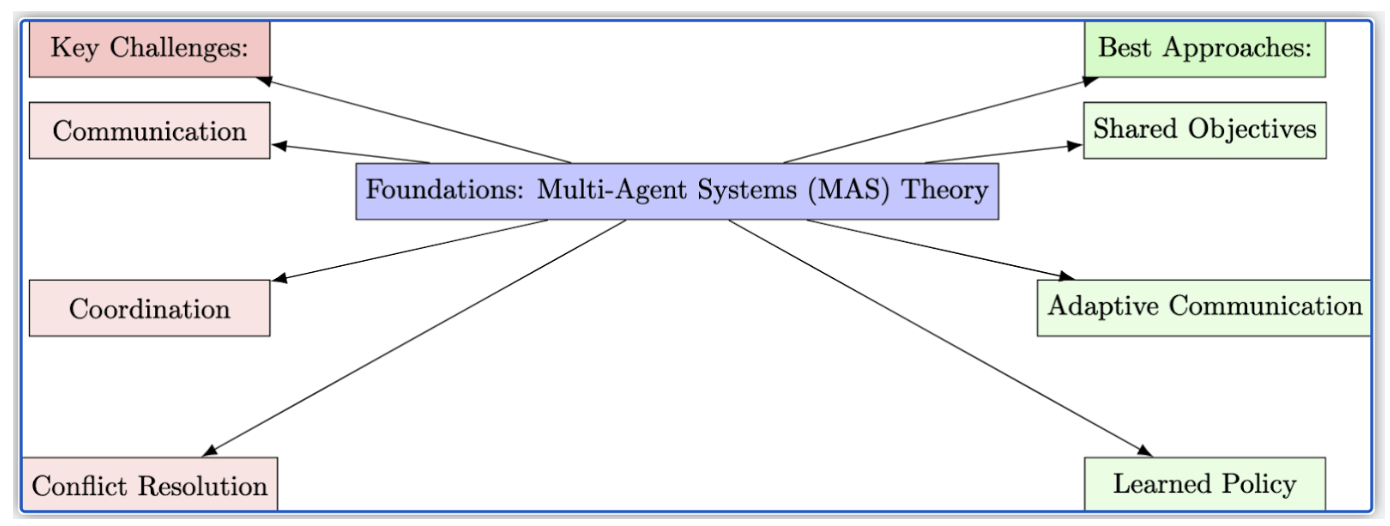}
    \caption{Using multi-agent systems theory to examine the key challenges in collaboration between embodied AI agents, and identifying the best approaches to tackle them.}
    \label{fig:enter-label}
\end{figure}

\textbf{Social Intelligence and Human-Agent Interactions for Embodied Agents.}
While multi-agent collaboration focuses on the interaction between AI agents, social intelligence emphasizes the interaction between AI agents and humans, which is equally vital for the success of embodied AI systems. Social intelligence is an important aspect of embodied agents, as it can enable them to interact with humans in a more natural and effective way. It involves the ability to understand and interpret human behavior, emotions, and intentions, as well as to generate responses that are context-dependent and socially aware. 

Some key components of social intelligence in embodied agents include:

\begin{itemize}
    \item Emotion recognition enables agents to recognize and interpret human emotions, such as facial expressions, body language, and tone of voice.
    \item Empathy and understanding enables agents to understand the feelings of users, and to respond in a way that is sensitive to their emotional state.
    \item Social reasoning enables agents to reason about social situations and interactions, including understanding social norms, roles, and relationships.
    \item Communication and interaction enables agents to communicate effectively with humans, using verbal and non-verbal cues, and to engage in meaningful interactions that are context-dependent and socially aware.
\end{itemize}

\section{Ethical Considerations of Embodied AI Agents}
\label{sec:ethics}
As embodied AI agents interact with users in both digital and physical spaces, they pose significant ethical risks that can compromise user trust and well-being. Two pressing concerns are the protection of user data and the potential for anthropomorphism, where users may be more susceptible to biases, hallucinations, or misinformation due to their increased trust in these agents. Regulatory bodies will likely prioritize mitigating these risks to ensure the safe and responsible development of embodied AI agents.

\paragraph{Privacy and Security } Unlike many traditional AI models—such as chatbots or coding assistants—that operate in isolated or task-specific environments, embodied agents are integrated into the very fabric of our daily lives. They inhabit the physical world alongside us, observing, learning, and adapting in real time. Consider, for example, an AI agent  embedded in a wearable device, such as smart glasses. It can potentially listen to our conversations, accompany us anywhere we go, see what we see, and hear what we hear. Similarly, robotic agents positioned within our homes have intimate access to our private spaces and routines, and develop a deep contextual awareness of our behaviors, preferences, and environments. This level of embodiment opens up unprecedented opportunities for memory, learning and adaptation. However, this powerful capability also comes with potential privacy implications. The data these agents are exposed to is not only vast but also sensitive, ranging from personal conversations and habits to potentially personal or private information. To function effectively, embodied agents might require storing some of this data, either directly or indirectly through updated model weights. The data stored and the model weights should therefore be protected. One simple way to do this is to store model weights and personal data on-device in an encrypted form; this protects against an adversary who has physical access to the device. 

A second challenge is that we might want to use some of this data to update and improve centralized models over time. A technical solution to this is federated learning \citep{kairouz2021advances,li2020federated}, where data stays on-device, and only gradients are transmitted to a central server for updating the model in question. It is also well known that federated learning by itself does not offer sufficient privacy \citep{geiping2020inverting}, as gradients during training can be reversed; privacy can be obtained by combining with a privacy-preserving technique such as differential privacy \citep{dwork2006calibrating}, but this might lead to loss of accuracy \citep{chaudhuri2011differentially,yu2024vip}. 

Another challenge is whether embodied agents know what is sensitive and use sensitive data appropriately in their day to day decision making - in particular, whether they are properly doing data minimization during their operation. Recent work has shown that this is not always a given in the context of web-navigation agents \citep{zharmagambetov2025agentdam}, and that these agents sometimes leak sensitive information that is not strictly necessary to complete a given task. How to prevent agents from doing so and obeying privacy norms is an active area of research. Building highly performant agents that protect privacy is a priority for responsible developers.

\paragraph{Anthropomorphism } Anthropomorphism, the attribution of human characteristics or behavior to non-human entities \citep{epley2007seeing}, has been a topic of interest in the field of Human-Computer Interaction (HCI) and Artificial Intelligence (AI). While anthropomorphic design can enhance user engagement and experience \citep{coronado2022evaluating}, it also raises several technical and ethical concerns.

%\textbf{Overestimation of Capabilities: The Problem of Illusory Agency}
When AI agents are designed to mimic human-like behavior, users may overestimate their capabilities, leading to unrealistic expectations \citep{nass2000machines}. This phenomenon is known as illusory agency, where users attribute human-like intentions and abilities to the AI agent. As a result, users may become disappointed, frustrated, or even experience safety issues if the agent is unable to perform tasks that were assumed to be within its capabilities.
%\textbf{Lack of Transparency: The Need for Explainability}
Anthropomorphic design can obscure the fact that AI agents are machines, making it difficult for users to understand their limitations and decision-making processes. This lack of transparency can erode trust and make it challenging to identify and address potential biases or errors. To mitigate this issue, designers must adopt a user-centered approach that prioritizes transparency, autonomy, and respect for user values \citep{shneiderman2010designing}.
%\textbf{Social Manipulation: The Risk of Persuasive Design}
AI agents that exhibit human-like behavior could influence users real world behaviour and responses. For instance, if an AI-powered chatbot recommends that a user take a particular action or uses emotional language, a user’s behaviour might be shaped by their interactions with the chatbot, potentially even acting in a way they might not otherwise have done. To mitigate this risk, designers must adopt a responsible and transparent approach to AI development, and it may also become important to educate people about how to interact with, and act upon interactions with AI agents. 

%\textbf{Dependence on Human-Like Interaction: The Need for Alternative Interfaces}
Over-reliance on anthropomorphic design can create a dependence on human-like interaction, which may not always be possible or desirable. In situations where the AI agent is unable to respond in a human-like manner, users may become frustrated or disengage from the interaction altogether. To address this issue, researchers have proposed alternative interface designs, such as voice-based interfaces or gesture-based interfaces \citep{oviatt2007multimodal}.

%\textbf{Designing Responsible AI Agents: }
To mitigate ethical issues associated with anthropomorphism, we propose to adopt a responsible and transparent approach to embodied AI agent development. This involves clearly communicating the agent's capabilities and limitations and prioritizing responsible design patterns.

By acknowledging the potential pitfalls of anthropomorphism and designing responsible AI agents, we can create more effective, trustworthy, and user-centered interactions. Other than the ethical challenges of privacy, security and anthropomorphism, other challenges remain at the underlying foundational model level, which includes human value alignment \citep{cahyawijaya2024high}, model biases and discrimination, and of course, hallucination \citep{bang2025hallulens}. To address these challenges, researchers at Meta are actively working on identifying and mitigating biases \citep{qian2022perturbation,esiobu2023robbie}, identifying and mitigating hallucinations \citep{ji2025calibrating}, as well as human value alignment \citep{zhou2023lima,yu2024robust}.

\section{Conclusion}
The exploration of embodied AI agents highlights their transformative potential across various domains, including virtual, wearable, and robotic applications. These agents, by integrating perception and action, offer a more human-like interaction with both digital and physical environments. The development of world models is crucial, enabling these agents to understand and predict their surroundings, thus enhancing their ability to perform complex tasks autonomously. Virtual embodied agents, such as AI avatars, are revolutionizing fields like therapy and entertainment by providing emotionally intelligent interactions.They need to be able to learn and act in digital and simulated worlds. Wearable agents, exemplified by smartglasses, offer real-time assistance and personalized experiences but they need to understand the physical world better and plan with reasoning. Robotic agents promise to alleviate labor shortages and perform tasks in unstructured environments.

We also emphasize the importance of ethical considerations, particularly regarding privacy/security and anthropomorphism, as these agents become more integrated into daily life. Future research directions include improving the both learning and action of these agents, enhancing their social intelligence and multi-agent collaboration, and ensuring ethical standards are met. By addressing these challenges, embodied AI agents hold the promise of transforming human-technology interaction, making it more intuitive and responsive to human needs. The ongoing advancements in this field will continue to push the boundaries of AI capabilities, paving the way for a future where AI seamlessly integrates into our lives.

\clearpage
\newpage
\bibliographystyle{assets/plainnat}
\bibliography{paper}

\end{document}